\RequirePackage{fix-cm}
\documentclass[twocolumn]{svjour3}          % twocolumn
\smartqed  % flush right qed marks, e.g. at end of proof
\usepackage{graphicx}
\usepackage{booktabs}

\usepackage{tabularx}
\usepackage{marvosym}
\usepackage{subfig}
\usepackage{algorithm}
\usepackage{algpseudocode}
\usepackage{fixltx2e}
\MakeRobust{\Call}
\usepackage{hyperref}
\hypersetup{colorlinks=true,linkcolor=black,citecolor=black,filecolor=black,urlcolor=black}
\usepackage[square,sort,comma,numbers]{natbib}
\newcommand{\envelope}{(\raisebox{-.5pt}{\scalebox{1.45}{\Letter}}\kern-1.7pt\,)}
\usepackage{microtype}

\journalname{Evol. Intel.}
\begin{document}

\title{A semantic network-based evolutionary algorithm for computational creativity}



\author{At{\i}l{\i}m~G\"{u}ne\c{s}~Baydin \and
        Ramon~L\'{o}pez~de~M\'{a}ntaras \and
        Santiago Onta\~{n}\'{o}n
}

\authorrunning{A. G. Baydin, R. L\'{o}pez~de~M\'{a}ntaras, S. Onta\~{n}\'{o}n} % if too long for running head

\institute{At{\i}l{\i}m~G\"{u}ne\c{s}~Baydin \envelope\at
                Hamilton Institute \& Department of Computer Science\\
                National University of Ireland Maynooth\\
                Co. Kildare, Ireland\\
                \email{atilimgunes.baydin@nuim.ie}
            \and
            Ramon~L\'{o}pez~de~M\'{a}ntaras \at
                Artificial Intelligence Research Institute, IIIA\,-\,CSIC\\
                Campus Universitat Aut\`{o}noma de Barcelona\\
                08193 Bellaterra, Spain\\
                \email{mantaras@iiia.csic.es}
            \and
            Santiago Onta\~{n}\'{o}n \at
                Department of Computer Science\\
                Drexel University\\
                3141 Chestnut Street, Philadelphia, PA 19104, USA\\
                \email{santi@cs.drexel.edu}
}

\date{Received: date / Accepted: date}

\maketitle

\begin{abstract}

We introduce a novel evolutionary algorithm (EA) with a semantic network-based representation. For enabling this, we establish new formulations of EA variation operators, crossover and mutation, that we adapt to work on semantic networks. The algorithm employs commonsense reasoning to ensure all operations preserve the meaningfulness of the networks, using ConceptNet and WordNet knowledge bases. The algorithm can be interpreted as a novel memetic algorithm (MA), given that (1) individuals represent pieces of information that undergo evolution, as in the original sense of memetics as it was introduced by Dawkins; and (2) this is different from existing MA, where the word ``memetic'' has been used as a synonym for local refinement after global optimization. For evaluating the approach, we introduce an analogical similarity-based fitness measure that is computed through structure mapping. This setup enables the open-ended generation of networks analogous to a given base network.

\end{abstract}

\section{Introduction}
\label{SectionIntroduction}
%\paragraph{Paragraph headings} Use paragraph headings as needed.

We introduce an evolutionary algorithm (EA) that generates semantic networks under a fitness measure based on information content and structure. This algorithm is, to the best of our knowledge, the first instance in literature where semantic networks are created via an evolutionary optimization process and specially developed structural variation operators respect the semantics of commonsense relations.

The algorithm works by fitness-based selection and reproduction of networks undergoing gradual changes introduced by variation operators. The initial generation of networks, and the variation operators of mutation and crossover, make use of randomly picked concepts and relations that are associated with existing nodes in a network, queried from commonsense knowledge bases. We currently use ConceptNet and WordNet knowledge bases for this purpose. The gradual changes in the algorithm are thus driven by randomness constrained by commonsense knowledge.

We demonstrate the approach via a fitness function measuring analogical similarity to a given base network. This is particularly interesting from an analogical reasoning perspective, because it enables us to spontaneously generate analogical mappings \emph{and} novel analogous cases, in contrast with existing algorithms capable of generating \emph{only} the mapping between two given cases. Spontaneously generating novel networks that are analogous to a given network, this demonstration is relevant for computational creativity applications, where methods simulating analogical creativity are sought for tasks such as story generation.

Seeing the evolutionary optimization of information represented within semantic networks as an implementation of the idea of ``memes'' in cultural evolution, this algorithm can be interpreted as a novel type of memetic algorithm (MA). In this designation, we use the term ``memetic'' in a different technical sense from existing models classified as MA, and with an implication closer to the original meaning as it was first introduced as a metaphor by \citet{Dawkins1976} in his book \emph{The Selfish Gene} and later popularized by \citet{Hofstadter1981}.

This is due to several reasons.

Within the existing field of MA, one models the effects of cultural evolution as a local refinement process for each individual, running on top of a global, population-based, optimization \citep{Moscato2004}. So, the emphasis is on the local refinement of each individual due to memetic evolutionary factors\footnote{From a biological perspective, this sense emphasizes the effect of society, culture, and learning on the survival of individuals on top of their physical traits emerging through genetic evolution. An example would be the use of knowledge and technology by the human species to survive in diverse environments, far beyond the physical capabilities available to them solely by the human anatomy.}. In algorithmic terms, this results in a combination of population-based global search with a local search step run for each individual. Thus, the only connection of the existing work in MA with the idea of ``memetics'' is using this word as a synonym for ``local refinement of candidate solutions''.

In contrast, the emphasis in our approach is directly on the memetic evolution itself, given

\begin{enumerate}
    \item it is the units of information (represented as semantic networks) that are undergoing variation, reproduction, and selection, exactly as in the original metaphor by \citet{Dawkins1976}; 
    \item we have variation operators developed specific for this knowledge representation-based approach, respecting the semantics and commonsensical correctness of the evolving structures; and
    \item the whole process is guided by a fitness measure that is defined as a function of some selected set of features of the knowledge represented by each individual.
\end{enumerate}

The article is organized as follows. In Sect.~\ref{SectionBackground} we provide background information on the subjects of evolution, creativity, and culture, followed by a brief review existing models of graph-based EA, to enable a discussion of how our contribution is related with existing work in the field. In Sect.~\ref{SectionTheAlgorithm}, we go over a detailed description of our algorithm, including details of representation and the newly introduced variation operators specific for semantic networks. We introduce the analogical similarity-based fitness measure in Sect.~\ref{SectionAnalogyAsAFitnessMeasure}, presenting results of experiments with the spontaneous generation of analogies. Sect.~\ref{SectionConclusions} ends the article with concluding remarks and a discussion of limitations and future directions for our approach.

\section{Background}
\label{SectionBackground}

\subsection{Evolution, creativity, and culture}

Following the success and explanatory power of evolutionary theory in biology, insights about the ubiquity of evolutionary phenomena have paved the way towards an understanding that these processes are not necessarily confined to biology. That is to say, whenever one has a system capable of exhibiting a kind of variation, heredity, and selection, one can formulate an evolutionary account of the complexity observed in almost any scale and domain. This approach is termed \emph{Universal Darwinism}, generalizing the mechanisms and extending the domain of evolutionary processes to systems outside biology, including economics, psychology, physics, and even culture \cite{Dennett1995,Bickhard2003}.

Within this larger framework, the concept of \emph{meme} first introduced as a metaphor by \citet{Dawkins1976} as an evolving unit of culture\footnote{Or, information, idea, or belief.} analogous to a \emph{gene}, hosted, altered, and reproduced in minds, later formed the basis of the approach called memetics\footnote{Quoting Dawkins \cite{Dawkins1976}: \emph{``Examples of memes are tunes, ideas, catch-phrases, clothes fashions, ways of making pots or of building arches. Just as genes propagate themselves in the gene pool by leaping from body to body via sperms or eggs, so memes propagate themselves in the meme pool by leaping from brain to brain...''}}.

Popularized by \citet{Hofstadter1981}, the explanation found itself use in cultural and sociological studies. For example, \citet{Balkin1998} argues that ideologies can be explained using a meme-based description, produced through processes of cultural evolution and transcending the lives of individuals. This evolutionary approach to ideology also enables genetic-inspired descriptions of cultural phenomena, such as \emph{ideological drift}. Similarly, creativity, as an integral part of culture, has also been addressed by these studies at the intersection of evolution and culture \citep{Gabora2010}. At this point, one has to clarify in which of the two highly related, but conceptually different, ways one uses the notions of ``evolution'' and ``creativity'' together.

The first point of view basically discusses the role of arts and creativity in the general framework of classical evolutionary biology, considering the provided advantages for adaptation and survival. All known societies enjoy creative pursuits such as literature, music, and visual arts; and there is evidence from the field of archaeology that this interest arose relatively early in the development of human species. Combined with the knowledge that a sense of aesthetic is also encountered in several other species of the animal kingdom, there is ample evidence to consider an ``evolutionary basis'' for creativity \cite{Martindale2007}.

Alternatively, inspired by the insight that evolutionary processes are not confined to biology, and using evolutionary theories of sociocultural change, one can consider that culture itself is possibly recreated through evolutionary processes occurring in the abstract environment of thoughts, concepts, or ideas. An example for this kind of interpretation in social sciences is the evolutionary epistemology theory put forth by Campbell \citep{Bickhard2003}.

Surely, a unifying approach considering all types of evolution is also possible, studying it as a general phenomenon applying at different levels to both physical and cultural systems. Within the creativity field, this kind of approach is taken by \citet{Skusa2002}, who study the processes occurring in systems exhibiting biological or cultural evolution. They base their insights on their work on evolutionary activity statistics for visualizing and measuring diverse systems \cite{Bedau1997}. Similarly, \citet{Gabora2010} investigate the issues at the intersection of creativity and evolution, from both biological and cultural senses.

A similar dichotomy also exists in the interpretation of the role of evolutionary algorithms in computational creativity.

Researchers realize that EA can be applied to computational creativity problems, considering them as a new area of complex and difficult technical problems where they can employ the proven power of EA as a black-box optimization tool.

Again, as in the case of sociocultural evolution, one can also consider the creativity process itself as taking place through evolutionary processes in an abstract ``creativity space''. In cases where evidence for an underlying evolutionary process can be spotted (as in the case of cultural evolution), in addition to providing solutions to difficult problems, one can also consider the simple but powerful explanatory power of evolutionary theory for the seemingly complex task of creativity.

The work that we present in this article is open to both interpretations. In addition to being a technique for the generation of semantic networks for a given creativity task, we can also use it---due to its memetic interpretation---to model the evolution of human culture through passing generations.

\subsection{Graph-based evolutionary algorithms}

There are several existing algorithms using graph-based representations for the encoding of candidate solutions in EA \cite{Montes2004}. The most notable work among these is genetic programming (GP) \cite{Koza2003}, where candidate solutions are pieces of computer program represented in a tree hierarchy. The trees are formed by \emph{functions} and \emph{terminals}, where the terminal set consists of variables and constants, and the function set can contain mathematical functions, logical functions, or functions controlling program flow, specific to the target problem.

In parallel distributed genetic programming (PDGP) \cite{Poli1999}, the restrictions of the tree structure of GP are relaxed by allowing multiple outputs from a node, which allows a high degree of parallelism in the evolved programs. In evolutionary graph generation (EGG) \cite{Chen2002} the focus is on evolving graphs with applications in electronic circuit design. Genetic network programming (GNP) \cite{Mabu2007} introduces compact networks with conditional branching and action nodes; and similarly, neural programming (NP) \cite{Teller1998} combines GP with artificial neural networks for the discovery of network structures via evolution.

The use of a graph-based representation makes the design of variation operators specific to graphs necessary.

In works such as GNP, this is facilitated by using a string-based encoding of node names, types, and connectivity, permitting operators very close to their counterparts in conventional EA; and in PDGP, the operations are simplified by making nodes occupy points in a fixed-size two-dimensional grid.

Our approach in this article, on the other hand, is closely related with how GP handles variation.

In GP crossover operation, two candidate solutions are combined to form two new solutions as their offspring. This is accomplished by randomly selecting \emph{crossover fragments} in both parents, deleting the selected fragment of the first parent and inserting the fragment from the second parent. The second offspring is produced by the same operation in reverse order. An important advantage of GP is its ability to create nonidentical offspring even in the case where the same parent is selected to mate with itself in crossover \footnote{This is in stark contrast with approaches such as GA, where a crossover operation of identical parents would yield identical offspring due to the linear nature of the representation}.

In GP, there are two main types of mutations: the first one involves the random change of the type of a function or terminal at a randomly selected position in the candidate solution; while in the second one an entire subtree of the candidate solution can be replaced by a new randomly created subtree.

What is common within GP related algorithms is that the output of each node in the graph can constitute an input to another node. In comparison, for the semantic network-based representation that we will introduce, the range of connections that can form a graph of a given set of concepts is constrained by commonsense knowledge, that is to say, the relations have to make sense to be useful. To address this issue, we introduce new crossover and mutation operations for memetic variation, making use of commonsense reasoning \cite{Mueller2006,Havasi2007} and adapted to work on semantic networks.

Of the existing graph-based EAs, the implementation nicknamed McGonagall by \citet{Manurung2003} bears similarities to our approach in that it uses a ``flat semantic representation'' that is essentially equivalent to what we here call semantic networks. McGonagall uses an EA approach to poetry generation, using fitness measures involving poetic metre evaluations and semantic similarity to a given target poem. The system uses special rule-based variation operators that ensure grammaticality and meaningfulness by exploiting domain knowledge. This is comparable to our use of commonsense reasoning to constrain variation operators to ensure meaningful semantic networks.

\section{The algorithm}
\label{SectionTheAlgorithm}

Our algorithm, outlined in Algorithm~\ref{AlgorithmNewMA}, proceeds similar to conventional EA, with a relatively small set of parameters. 

\begin{algorithm}
    \caption{Procedure for the novel semantic network-based memetic algorithm. Refer to Table~\ref{TableNewMAParameters} for an overview of involved parameters.}
    \label{AlgorithmNewMA}
    \begin{algorithmic}[1]
        \Procedure{MemeticAlgorithm}{}
            \State $P(t=0) \gets$ \Call{InitializePopulation}{$Size_{pop}$, $Size_{network}$, $Score_{min}$, $Count_{timeout}$}
            \Repeat
                \State $\phi(t) \gets$ \Call{EvaluateFitnesses}{$P(t)$}
                \State $N(t) \gets$ \Call{NextGeneration}{$P(t)$, $\phi(t)$, $Size_{pop}$, $Size_{tourn}$, $Prob_{win}$, $Prob_{rec}$, $Prob_{mut}$, $Score_{min}$, $Count_{timeout}$}
                \State $P(t+1) \gets N(t)$
                \State $t \gets t+1$
            \Until{stop criterion}
        \EndProcedure
    \end{algorithmic}
\end{algorithm}

\begin{table*}
    \caption{Parameter set of the novel evolutionary algorithm (Algorithm~\ref{AlgorithmNewMA}).}
    \label{TableNewMAParameters}
    \begin{tabularx}{\textwidth}{@{}p{1.8cm}llX@{}}
        \toprule
        & Parameter & Interval & Explanation\\
        \midrule
        Evolution & $Size_{pop}$ & $[1, \infty)$ & Number of individuals forming the population\\
        & $Prob_{rec}$ & $[0, 1]$ & Probability of applying crossover operation\\
        & $Prob_{mut}$ & $[0, 1]$ & Probability of applying mutation operation\\
        \midrule
        Semantic & $Size_{network}$ & $[1, \infty)$ & Maximum size of randomly created semantic networks in the initial population\\
        networks & $Score_{min}$ & $[-10, 10]$ & Minimum quality score of commonsense relations throughout the algorithm\\
        & $Count_{timeout}$ & $[1, \infty)$ & Timeout value for the number of trials in commonsense retrieval operations\\
        \midrule
        Tournament selection & $Size_{tourn}$ & $[1, Size_{pop}]$ & Number of individuals randomly selected from the population for each tournament selection event\\
        & $Prob_{win}$ & $[0, 1]$ & The probability that the best individual in the ranked list of tournament participants wins the tournament\\
        \bottomrule
    \end{tabularx}
\end{table*}

Descriptions of initialization, fitness evaluation, selection, and memetic variation steps are presented in detail in the following sections. The parameters affecting each step of the algorithm, along with their explanations, are summarized in Table~\ref{TableNewMAParameters}.

\subsection{Semantic networks}

Semantic networks are graphs that represent semantic relations between concepts. In a semantic network, knowledge is expressed in the form of directed binary relations, represented by edges, and concepts, represented by nodes. This type of graph representation has found use in many subfields of artificial intelligence, including natural language processing, machine translation, and information retrieval \citep{Sowa2000}.

\begin{figure}
  \centering
  \includegraphics[width=\columnwidth]{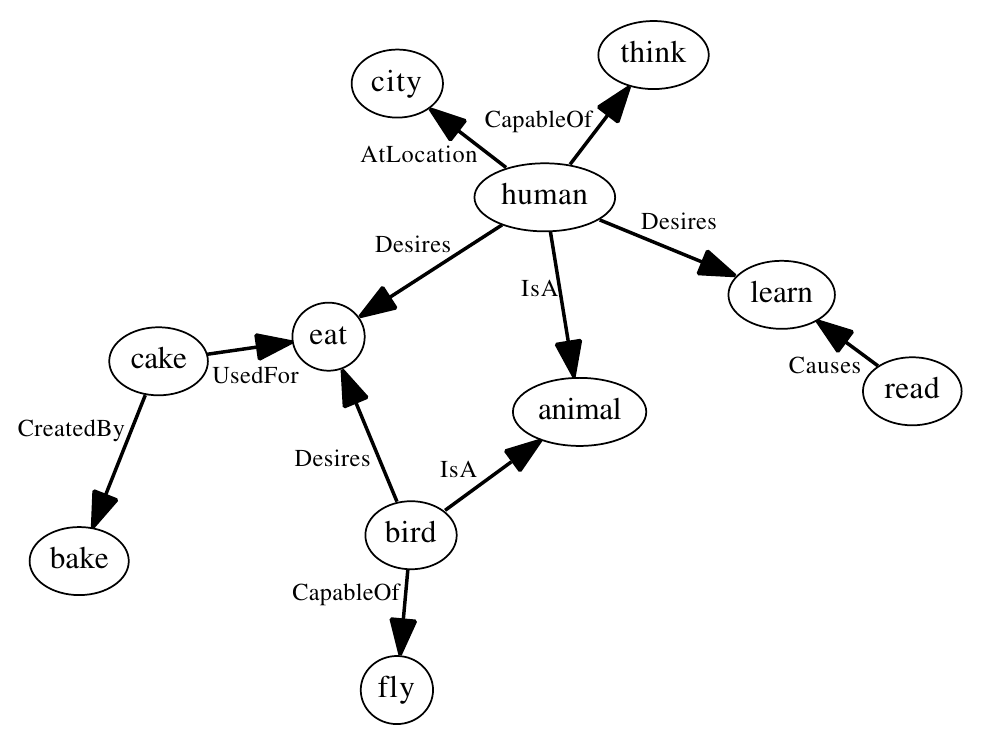}
  \caption{A semantic network with 11 concepts and 11 relations.}
  \label{FigureSemanticNetwork}
\end{figure}

Figure~\ref{FigureSemanticNetwork} shows a graph representation of a simple semantic network. In addition to the graphical representation, we also adopt the notation of \texttt{IsA(bird, animal)} to mean that the concepts \texttt{bird} and \texttt{animal} are connected by the directed relation \texttt{IsA($\cdot$,$\cdot$)}, i.e. ``bird is an animal''.

An important characteristic of a semantic network is whether it is definitional or assertional: in definitional networks the emphasis is on taxonomic relations (e.g. $IsA(human, mammal)$) describing a subsumption hierarchy that is true by definition; in assertional networks, the relations describe instantiations and assertions that are contingently true (e.g. $AtLocation(human, city)$) \cite{Sowa2000}. In this study, we combine the two approaches for increased expressivity. As such, semantic networks provide a simple yet powerful means to represent the ``meme'' metaphor of Dawkins as data structures that are algorithmically manipulatable, allowing a procedural implementation of memetic evolution.

\subsection{Commonsense reasoning}
\label{SubsectionCommonsenseReasoning}

A foundational issue that comes with our approach is the problem of reconciling the intrinsically random nature of evolutionary operations with the requirement that the evolving semantic networks should be meaningful. 

This is so because, unlike existing graph-based approaches such as GP or GNP, not every node in a semantic network graph can be connected to an arbitrary other node through an arbitrary type of relation. This issue is relevant in every type of modification operation that needs to be executed during the course of our algorithm.

Simply put, the operations should be constrained by commonsense knowledge: a relation such as \texttt{IsA(bird, animal)} is meaningful, while \texttt{Causes(bird, table)} is not. 

We address this problem by utilizing the nascent subfield of AI named \emph{commonsense reasoning} \citep{Mueller2006,Havasi2007}. Within AI, since the pioneering work by \citet{McCarthy1958}, commonsense reasoning has been commonly regarded as a key ability that a system must possess in order to be considered truly intelligent \citep{Minsky2006}. 

Commonsense reasoning refers to the type of reasoning involved in everyday human thinking, based on \emph{commonsense knowledge} that an ordinary person is expected to know, or ``the knowledge of how the world works'' \citep{Mueller2006}. It comprises information such as \texttt{HasA(human, brain)}, \texttt{IsA(sun, star)}, or \texttt{CapableOf(ball, roll)}, which are acquired and taken for granted by any adult human, but which need to be introduced in a particular way to a computational reasoning system.

Knowledge bases such as the Cyc project maintained by Cycorp company\footnote{\url{http://www.cyc.com/}}, ConceptNet project of MIT Media Lab\footnote{\url{http://conceptnet5.media.mit.edu/}}, and the Never-Ending Language Learning (NELL) project of Carnegie Mellon University\footnote{\url{http://rtw.ml.cmu.edu/rtw/}} are set up to collect and classify commonsense information for the use of research community. In our current implementation, we make use of ConceptNet and the lexical database WordNet\footnote{\url{http://wordnet.princeton.edu}} to address the restrictions of processing commonsense knowledge.

\subsubsection{Knowledge bases}

The ConceptNet project is a part of the Open Mind Common Sense (OMCS) initiative of the MIT Media Lab, based on the input of commonsense knowledge from general public through several ways, including parsed natural language and semi-structured fill-in-the-blanks type forms \citep{Havasi2007}. As of 2013, ConceptNet is in version 5 and, in addition to data collected through OMCS, it has been extended to include other data sources such as the Wikipedia and Wiktionary projects of the Wikimedia Foundation\footnote{\url{http://www.wikimedia.org/}} and the DBPedia project\footnote{\url{http://dbpedia.org/}} of the University of Leipzig and the Freie Universit\"{a}t Berlin. 

Access to the ConceptNet database is provided through a web API using JavaScript Object Notation (JSON) textual data format. Due to performance reasons, we use the previous version of ConceptNet, version 4, in our implementation. This is because of the high volume of queries to ConceptNet during the creation of random semantic networks and the application of variation operators. ConceptNet 4 provides the complete dataset in locally accessible and highly efficient SQLite database format, enabling substantially faster access to data compared with the web API and JSON format of the current version.

According to the study by \citet{Diochnos2013}, ConceptNet version 4 includes 566,094 assertions and 321,993 concepts. The variety of assertions in ConceptNet, initially contributed by volunteers from general public, makes it somewhat prone to noise. According to our experience, noise is generally due to charged statements about political issues, biased views about gender issues, or attempts of making fun. We address the noise problem by ignoring all assertions with a reliability score (determined by contributors' voting) below a set minimum $Score_{min}$ (Table~\ref{TableNewMAParameters})\footnote{The default reliability score for a statement is 1 \citep{Havasi2007}; and zero or negative reliability scores are a good indication of information that can be considered noise.}.

The lexical database WordNet \citep{Fellbaum1998} maintained by the Cognitive Science Laboratory at Princeton University also has characteristics of a commonsense knowledge base that make it attractive for our purposes. WordNet is based on a grouping of words into \emph{synsets} or \emph{synonym rings} which hold together all elements that are considered semantically equivalent\footnote{Another definition of \emph{synset} is that it is a set of synonyms that are interchangeable without changing the truth value of any propositions in which they are embedded.}.

In addition to these synset groupings, WordNet includes \emph{pointers} that are used to represent relations between the words in different synsets. These include semantic pointers that represent relations between word meanings and lexical pointers that represent relations between word forms.

For treating WordNet as a commonsense knowledge base compatible with ConceptNet, we define the set of correspondences we outline in Table~\ref{TableWordNetConceptNetRelations}. Similar approaches have also been used by other researchers in the field, such as by \citet{Kuo2010}.

\begin{table}
  \caption{Set of correspondences we define between WordNet and ConceptNet relation types.}
  \label{TableWordNetConceptNetRelations}
  \begin{tabularx}{\columnwidth}{@{}lX|lX@{}} 
    \toprule
    \multicolumn{2}{c}{WordNet} & \multicolumn{2}{c}{ConceptNet}\\
    Relation & Example & Relation & Example\\
    \midrule
    Hypernym & \emph{canine} is a hypernym of \emph{dog} & \texttt{IsA} & \texttt{IsA(dog, canine)}\\
    \midrule
    Holonym & \emph{automobile} is a holonym of \emph{wheel} & \texttt{PartOf} & \texttt{PartOf(wheel, automobile)}\\
    \midrule
    Meronym & \emph{wheel} is a meronym of \emph{automobile} & \texttt{PartOf} & \texttt{PartOf(wheel, automobile)}\\
    \midrule
    Attribute & \emph{edible} is an attribute of \emph{pear} & \texttt{HasProperty} & \texttt{HasProperty( pear, edible)}\\
    \midrule
    Entailment & \emph{to sleep} is entailed by \emph{to snore} & \texttt{Causes} & \texttt{Causes(sleep, snore)}\\
    \bottomrule
  \end{tabularx}
\end{table}

In the implementation of our algorithm, we answer the various types of queries to commonsense knowledge bases (such as the \textsc{RandomConcept()} call in Algorithm~\ref{AlgorithmNewMARandomNetwork}) via ConceptNet or WordNet on a random basis. When the query is answered by information retrieved from WordNet, we return the information formatted in ConceptNet structure based on the correspondences outlined in Table~\ref{TableWordNetConceptNetRelations} and attach the maximum reliability score of 10, since the information in WordNet is provided by domain experts and virtually devoid of noise.

In our implementation we use WordNet version 3, contributing definitional relations involving around 117,000 synsets. Another thing to note here is that, in ConceptNet version 5, WordNet already constitutes one of the main incorporated data sources. This means that, in case we switch from ConceptNet version 4 to version 5, our approach of accessing WordNet would be obsolete.

\subsection{Initialization}

At the start of a run, the population of size $Size_{pop}$ is initialized (Algorithm~\ref{AlgorithmNewMAInitializePopulation}) with individuals created through a procedure that we call \emph{random semantic network generation} (Algorithm~\ref{AlgorithmNewMARandomNetwork}), capable of assembling random semantic networks of any given size. 

\begin{figure*}
  \centering
  \subfloat[]{\includegraphics[height=3.5cm]{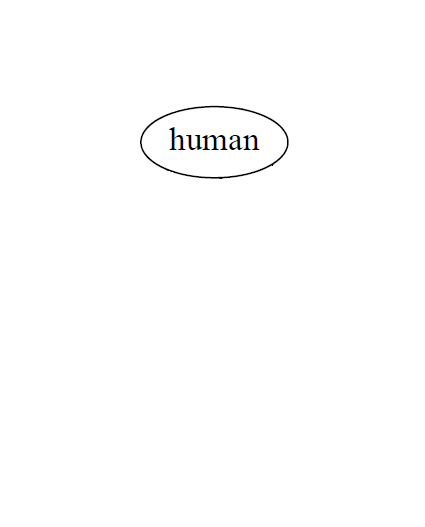}}
  \subfloat[]{\includegraphics[height=3.5cm]{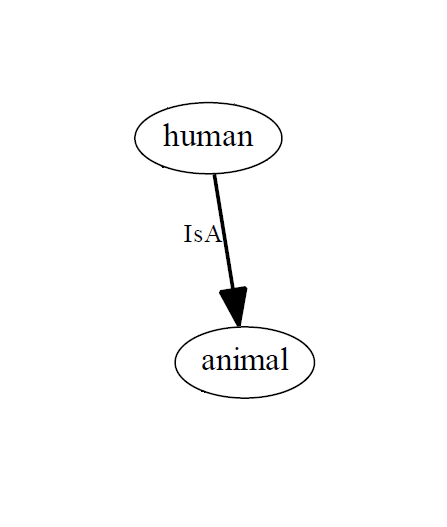}}
  \subfloat[]{\includegraphics[height=3.5cm]{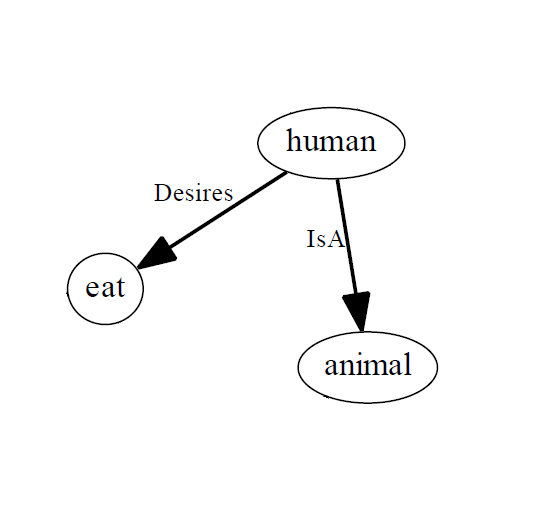}}
  \subfloat[]{\includegraphics[height=3.5cm]{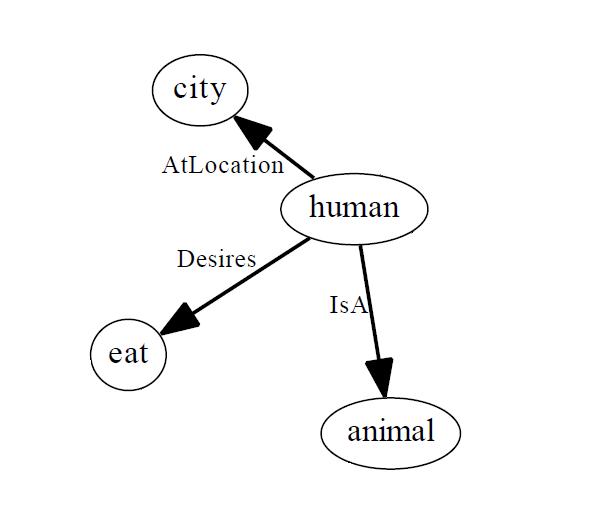}}
  \subfloat[]{\includegraphics[height=3.5cm]{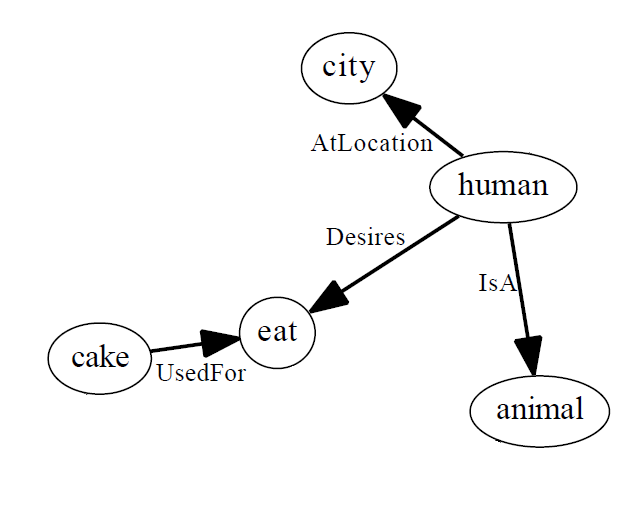}}
  \caption{The process of random semantic network generation, starting with a single random concept in (a) and proceeding with (b), (c), (d), (e), adding new random concepts from the set of concepts related to existing ones.}
  \label{FigureRandomNetworkGeneration}
\end{figure*}

Figure~\ref{FigureRandomNetworkGeneration} presents an example of a random semantic network created via this procedure. This works by starting from a network comprising a sole concept randomly picked from commonsense knowledge bases and running a semantic network expansion algorithm that

\begin{enumerate}
    \item randomly picks a concept in the given network (e.g. \texttt{human});
    \item compiles a list of relations, from commonsense knowledge bases, that the picked concept can be involved in (e.g. \texttt{CapableOf(human,} \texttt{think)}, \texttt{Desires(human,} \texttt{eat)}, \ldots);
    \item appends to the network a relation randomly picked from this list, together with the other involved concept; and 
    \item repeats this process until a given number of concepts have been appended to the network, or a set timeout $Count_{timeout}$ has been reached (as a failsafe for situations where there are not enough relations involving the concepts in the network being created).
\end{enumerate}

It is very important to note here that \emph{even if it is grown in a random manner, the generated network itself is totally meaningful, because it is a combination of meaningful pieces of information harvested from commonsense knowledge bases}.

The initialization algorithm depends upon the parameters of $Size_{network}$, the intended number of concepts in the randomly created semantic networks, and $Score_{min}$, the minimum ConceptNet relation score that should be satisfied by the retrieved relations (Table~\ref{TableNewMAParameters}).

\begin{algorithm}
    \caption{Procedure for the creation of initial random population.}
    \label{AlgorithmNewMAInitializePopulation}
    \begin{algorithmic}[1]
        \Procedure{InitializePopulation}{$Size_{pop}$, $Size_{network}$, $Score_{min}$, $Count_{timeout}$}
            \State \textbf{initialize} $P$\Comment{The return array}
            \For{$Size_{pop}$ times}
                \State $r \gets$ \Call{RandomNetwork}{$Size_{network}$, $Score_{min}$, $Count_{timeout}$}\Comment{Generate a new random network}
                \State \Call{AppendTo}{$P$, $r$}
            \EndFor
            \State \textbf{return} $P$
        \EndProcedure
    \end{algorithmic}
\end{algorithm}

\begin{algorithm}
    \caption{The random semantic network generation algorithm. The algorithm is presented here in a form simpler than the actual implementation, for the sake of clarity.}
    \label{AlgorithmNewMARandomNetwork}
    \begin{algorithmic}[1]
        \Procedure{RandomNetwork}{$Size_{network}$, $Score_{min}$, $Count_{timeout}$}
            \State \textbf{initialize} $net$\Comment{Empty return network}
            \State \textbf{initialize} $c$\Comment{Random initial seed concept}
            \For{$Count_{timeout}$ times}
                \State $c \gets$ \Call{RandomConcept}{$Score_{min}$}
                \State $rels \gets$ \Call{InvolvedRelations}{$c$}
                \If{\Call{Size}{$rels$} $\geq Size_{network}$}
                    \State \Call{AppendTo}{$net$, $c$}
                    \State \textbf{break for}\Comment{Favor a seed with more than a few relations}
                \EndIf
            \EndFor
            \State $t \gets 0$
            \Repeat
                \State $c \gets$ \Call{RandomConceptIn}{$net$}
                \State $rels \gets$ \Call{InvolvedRelations}{$c$}\Comment{The set of relations involving $c$}
                \State $r \gets$ \Call{RandomRelationIn}{$rels$}
                \If{\Call{Score}{$r$} $\geq Score_{min}$}
                    \State \Call{AppendTo}{$net$, $r$}\Comment{Append to the network $net$ the relation $r$ and its involved concepts}
                \EndIf
                \State $t \gets t + 1$
            \Until{\Call{Size}{$net$} $\geq Size_{network}$ or $t \geq Count_{timeout}$}
            \State \textbf{return} $net$
        \EndProcedure
    \end{algorithmic}
\end{algorithm}

\subsection{Fitness measure}

After the initial generation is populated by individuals created by the random semantic network generation algorithm that we outlined, the algorithm proceeds by assigning fitness values to each individual. Since our approach constitutes the first instance of semantic network-based EA, it falls on us to introduce fitness measures of interest for its validation.

As an example for showcasing our approach, in Sect.~\ref{SectionAnalogyAsAFitnessMeasure}, we define a fitness measure based on analogical similarity to an existing semantic network, giving rise to spontaneous generation of semantic networks that are in each generation more and more structurally analogous to a given network.

In general terms, a direct and very interesting application of our approach would be to devise realistically formed fitness functions modeling selectionist theories of knowledge, which remain untested until this time. One such theory is the \emph{evolutionary epistemology} theory of Campbell \citep{Bickhard2003}, which describes the development of human knowledge and creativity through selectionist principles, such as the \emph{blind variation and selective retention} (BVSR) principle.

It is also possible to make the inclusion of certain concepts in the evolving semantic networks a requirement, allowing the discovery of networks formed around a given set of seed concepts. This can be also achieved through starting the initialization procedure (Algorithm~\ref{AlgorithmNewMARandomNetwork}) with the given seed concepts.

After all the individuals in the current generation are assigned fitness values, the algorithm proceeds with the creation of the next generation of individuals through variation operators (Algorithm~\ref{AlgorithmNewMANextGeneration}). But before this, the algorithm has to apply selection to pick individuals from the current population that will be ``surviving'' to produce offspring.

\subsection{Selection}

After the assignment of fitness values, individuals are replaced with offspring generated via variation operators applied on selected parents. We employ tournament selection, because it is better at preserving population diversity\footnote{\emph{Diversity}, in EA, is a measure of homogeneity of the individuals in the population. A drop in diversity indicates an increased number of identical individuals, which is not desirable for the progress of evolution.} and allowing selection pressure to be adjusted through simple parameters \citep{Pohlheim2006}.

Tournament selection involves, for each selection event, running ``tournaments'' among a group of $Size_{tourn}$ randomly selected individuals. Individuals in the tournament pool then challenge each other in pairs and the individual with the higher fitness will win with probability $Prob_{win}$. This method simulates biological mating patterns in which two members of the same sex compete to mate with a third one of different sex for the recombination of genetic material. Individuals with higher fitness have better chance of being selected, but an individual with low fitness still has a chance, however small, to produce offspring. Adjusting parameters $Size_{tourn}$ and $Prob_{win}$ (Table~\ref{TableNewMAParameters}) gives us an intuitive and straightforward way to adjust the selection pressure on both strong and weak individuals.

In our implementation, we also allow reselection, meaning that the same individual from a particular generation can be selected more than once to produce offspring in different combinations. Algorithm~\ref{AlgorithmNewMASelect} gives an overview of the selection procedure that we implement.

\begin{algorithm}
    \caption{Implemented selection algorithm.}
    \label{AlgorithmNewMASelect}
    \begin{algorithmic}[1]
        \Procedure{Select}{$P(t)$, $\phi(t)$, $Size_{tourn}$, $Prob_{win}$}
            \State $w \gets$ \Call{RandomMember}{$P(t)$}\Comment{Current winner}
            \For{$Size_{tourn} - 1$ times}
                \State $o \gets$ \Call{RandomMember}{$P(t)$}\Comment{The next opponent}
                \If{\Call{LookupFitness}{$\phi(t)$, $o$} $\geq$ \Call{LookupFitness}{$\phi(t)$, $w$}}
                    \If{\Call{RandomReal}{0, 1} $\leq Prob_{win}$}
                        \State $w \gets o$\Comment{Opponent defeats current winner}
                    \EndIf
                \EndIf
            \EndFor
            \State \textbf{return} $w$
        \EndProcedure
    \end{algorithmic}
\end{algorithm}

\subsection{Memetic variation operators}

Variation operators form the last step in the cycle of our algorithm by creating the next generation of individuals before going back to the step of fitness evaluation (Algorithm~\ref{AlgorithmNewMA}).

As we mentioned in Sect.~\ref{SubsectionCommonsenseReasoning}, our representation does not permit arbitrary connections between different nodes in the network and requires special variation operators that should respect the commonsense structure of represented knowledge.

In the following sections, we present the \emph{commonsense crossover} and \emph{commonsense mutation} operators that we set up specific to semantic networks.

Using these operators, the next step in the cycle of our algorithm is the creation of the offspring through variation (Algorithm~\ref{AlgorithmNewMANextGeneration}). Crossover is applied to parents selected from the population until $Size_{pop} \times Prob_{rec}$ offspring are created (Table~\ref{TableNewMAParameters}), where each crossover event creates two offspring from two parents.

Following the tradition in the GP field \citep{Koza2003}, we design the variation process such that the offspring created by crossover do not undergo mutation. The mutation operator is applied only to the rest of individuals that are copied, or ``reproduced'', directly from the previous generation.

For generating the remaining part of the population, we reproduce $Size_{pop} \times (1 - Prob_{rec}) - 1$ number of individuals selected, and make these subject to mutation. We employ elitism: the last individual (hence the remaining $- 1$ in the previous equation) is a copy of the one with the current best fitness.

\begin{algorithm}
    \caption{Procedure for generating the next generation of individuals.}
    \label{AlgorithmNewMANextGeneration}
    \begin{algorithmic}[1]
        \Procedure{NextGeneration}{$P(t)$, $\phi(t)$, $Size_{pop}$, $Size_{tourn}$, $Prob_{win}$, $Prob_{rec}$, $Prob_{mut}$, $Score_{min}$, $Count_{timeout}$}
            \State \textbf{initialize} $N$\Comment{The return array}
            \State $c \gets {Size_{pop}} Prob_{rec} / 2$\Comment{Number of crossover events}
            \State $r \gets {Size_{pop}} - 2 c$\Comment{Number of reproduction events}
            \For{$c$ times}
                \State $p1 \gets$ \Call{Select}{$P(t)$, $\phi(t)$, $Size_{tourn}$, $Prob_{win}$}
                \State $p2 \gets$ \Call{Select}{$P(t)$, $\phi(t)$, $Size_{tourn}$, $Prob_{win}$}
                \State ${o1, o2} \gets$ \Call{Crossover}{$p1$, $p2$}\Comment{Crossover the two parents}
                \State \Call{AppendTo}{$N$, o1}\Comment{Two offspring from each crossover}
                \State \Call{AppendTo}{$N$, o2}
            \EndFor
            \For{$r - 1$ times}
                \State $m \gets$ \Call{Select}{$P(t)$, $\phi(t)$, $Size_{tourn}$, $Prob_{win}$}
                \State $m \gets$ \Call{Mutate}{$m$, $Prob_{mut}$}\Comment{Mutate an individual}
                \State \Call{AppendTo}{$N$, m}
            \EndFor
            \State \Call{AppendTo}{$N$, \Call{BestIndividual}{$P(t)$, $\phi(t)$}}\Comment{Elitism}
            \State \textbf{return} $N$
        \EndProcedure
    \end{algorithmic}
\end{algorithm}

\subsubsection{Commonsense crossover}

We introduce two types of \emph{commonsense crossover} that are tried in sequence by the variation algorithm.

The first type attempts a sub-graph interchange between two selected parents similar to common crossover in standard GP; and where this is not feasible due to the commonsense structure of relations forming the parents, the second type falls back to a combination of both parents into a new offspring.

\subsubsection{Type I Crossover (Subgraph Crossover)}

Firstly, a pair of concepts, one from each parent, that are \emph{interchangeable}\footnote{We define two concepts from different semantic networks as \emph{interchangeable} if both can replace the other in all, or part, of the relations the other is involved in, queried from commonsense knowledge bases.} are selected as \emph{crossover concepts}, picked randomly out of all possible such pairs.

\begin{figure*}
    \centering
    \subfloat[]{\includegraphics[width=7cm]{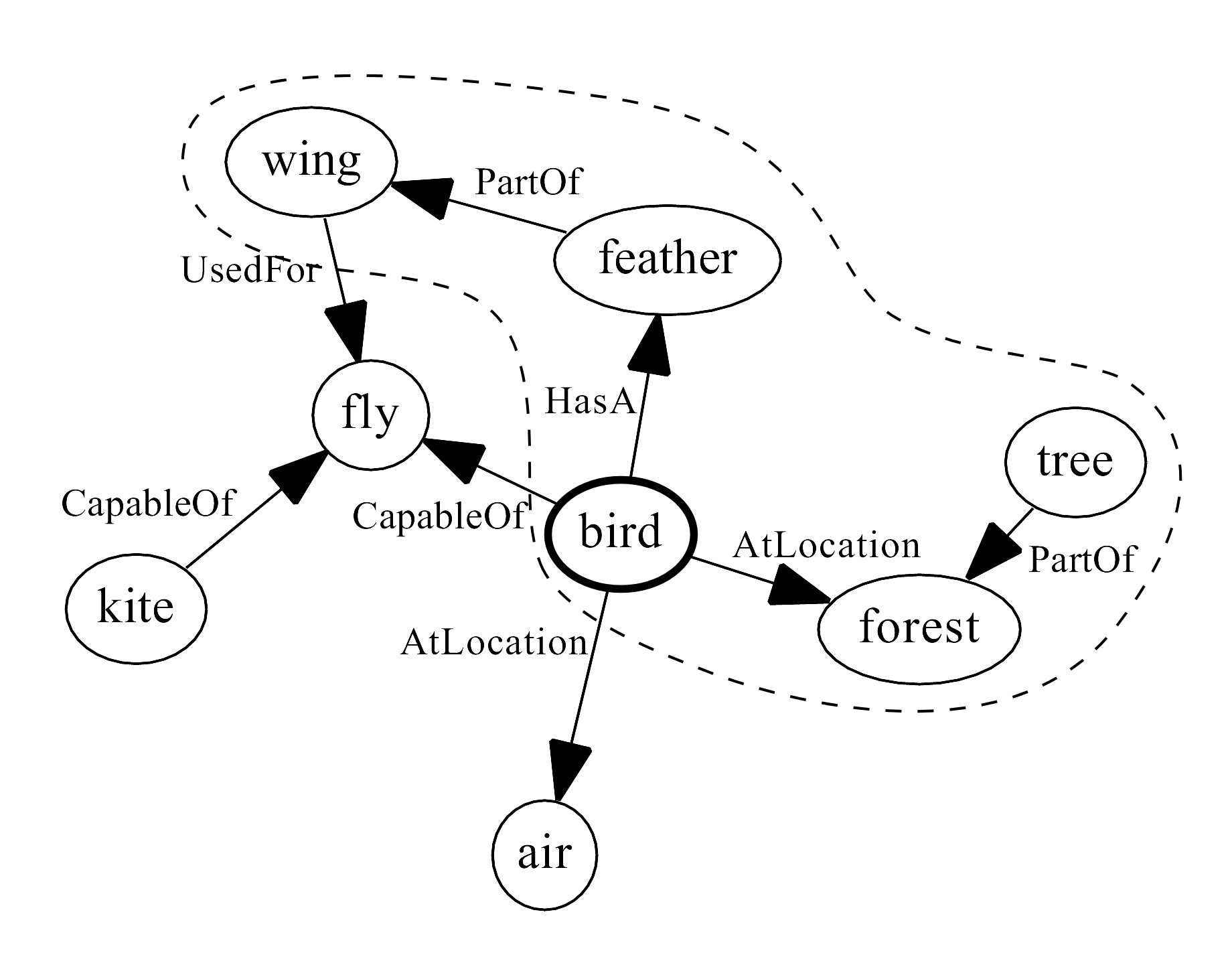}}
    \qquad
    \subfloat[]{\includegraphics[width=6.5cm]{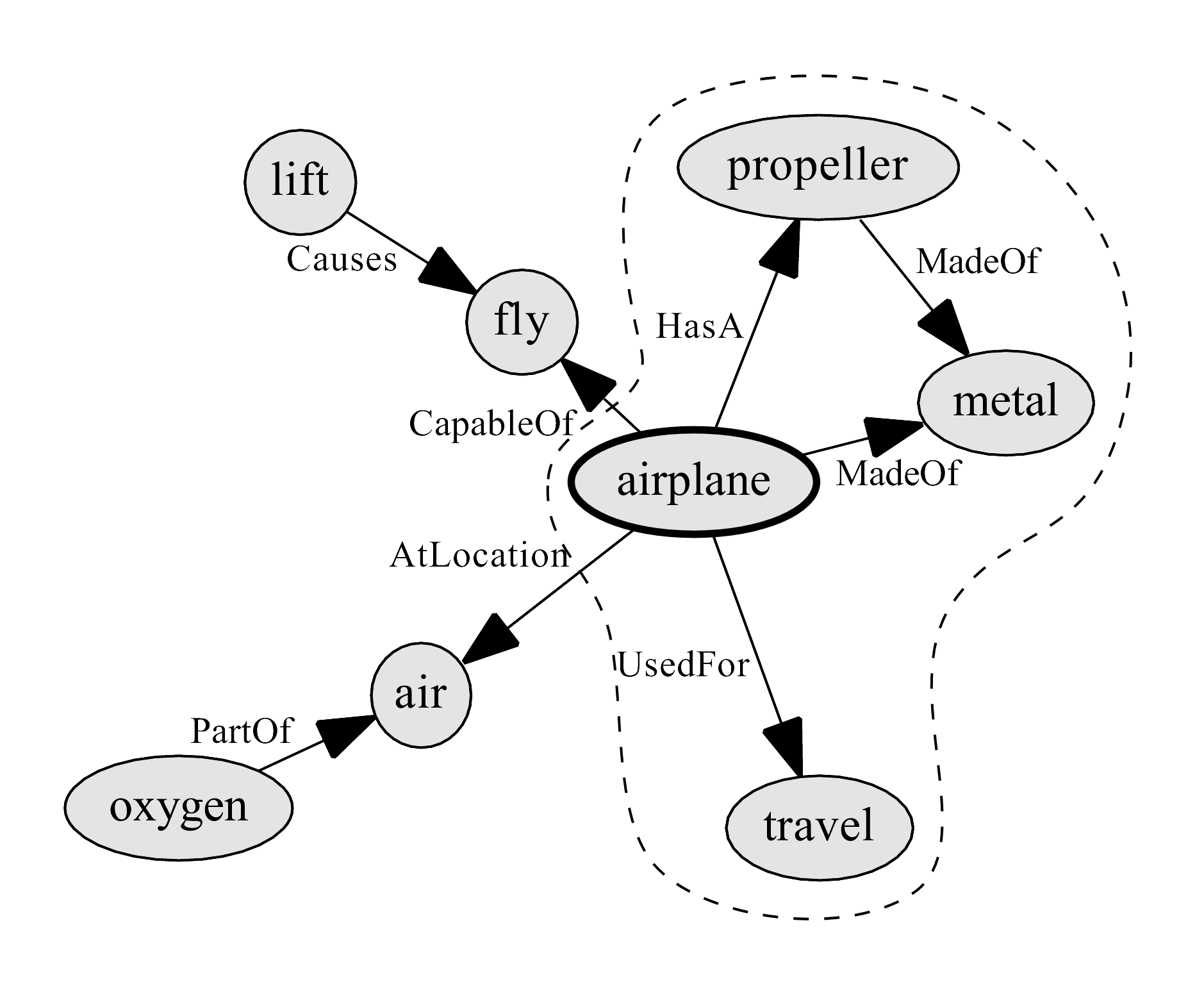}}
    \qquad
    \subfloat[]{\includegraphics[width=6cm]{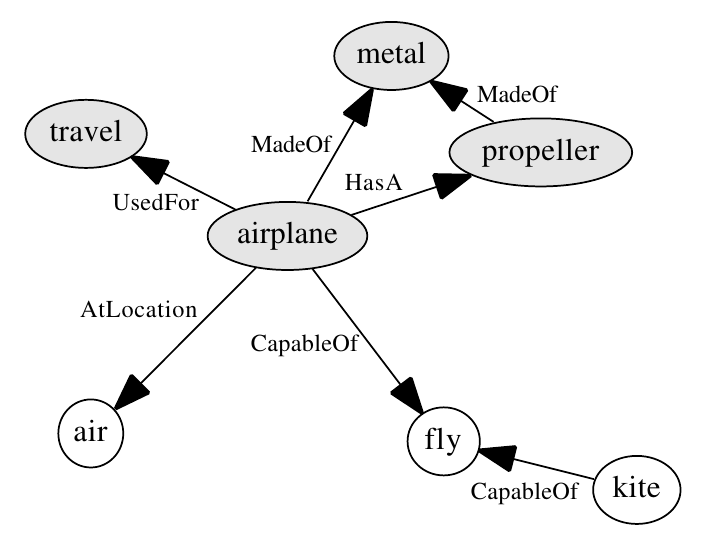}}
    \qquad
    \subfloat[]{\includegraphics[width=6.5cm]{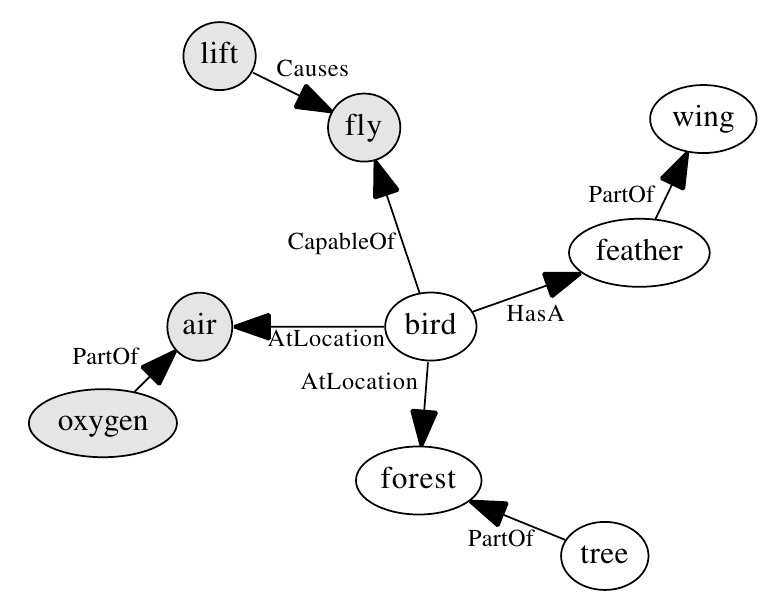}}
    \caption{Commonsense crossover type I (subgraph crossover). (a) Parent 1, centered on the concept \texttt{bird}; (b) Parent 2, centered on the concept \texttt{airplane}; (c) Offspring 1; (d) Offspring 2.}
    \label{FigureCrossoverTypeI}
\end{figure*}

For instance, for the parent networks in Figure~\ref{FigureCrossoverTypeI} (a) and (b), \texttt{bird} and \texttt{airplane} are interchangeable, since they can replace each other in the relations \texttt{CapableOf($\cdot$, fly)} and \texttt{AtLocation($\cdot$, air)}.

In each parent, a subgraph is formed, containing: 

\begin{enumerate}

\item the crossover concept;
\item the set of all relations, and associated concepts, that are not common with the other crossover concept 

\begin{sloppypar}
For example, in Figure~\ref{FigureCrossoverTypeI} (a), \texttt{HasA(bird, feather)} and \texttt{AtLocation(bird, forest)}; and in Figure~\ref{FigureCrossoverTypeI} (b), \texttt{HasA(airplane, propeller)}, \texttt{MadeOf(airplane, metal)}, and \texttt{UsedFor(airplane, travel)}; and 
\end{sloppypar}

\item the set of all relations and concepts connected to those found in the previous step, excluding the ones that are also one of those common with the other crossover concept.

\begin{sloppypar}
For example, in Figure~\ref{FigureCrossoverTypeI} (a) including \texttt{PartOf(feather, wing)} and \texttt{PartOf(tree, forest)}; and in Figure~\ref{FigureCrossoverTypeI} (b), including \texttt{MadeOf(propeller, metal)}); but excluding the concept \texttt{fly} in Figure~\ref{FigureCrossoverTypeI} (a), because of the relation \texttt{CapableOf($\cdot$, fly)}.
\end{sloppypar}

\end{enumerate}

This, in effect, forms a subgraph of information specific to the crossover concept, which is insertable into the other parent. Any relations between the subgraph and the rest of the network not going through the crossover concept are severed (e.g. \texttt{UsedFor(wing, fly)} in Figure~\ref{FigureCrossoverTypeI} (a)).

The two offspring are formed by exchanging these subgraphs between the parent networks (Figure~\ref{FigureCrossoverTypeI} (c) and (d)).

\subsubsection{Type {II} Crossover (Graph Merging Crossover)}

\begin{figure*}
    \centering
    \subfloat[]{\includegraphics[width=5cm]{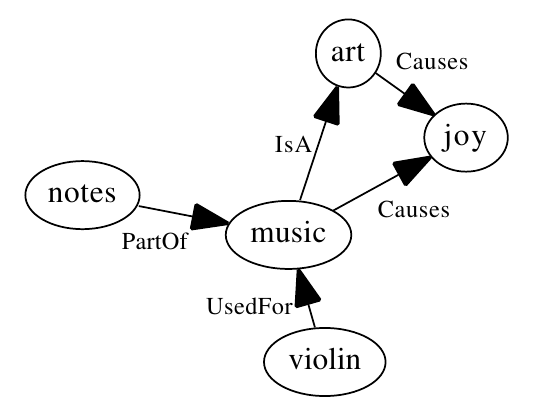}}
    \qquad
    \subfloat[]{\includegraphics[width=5cm]{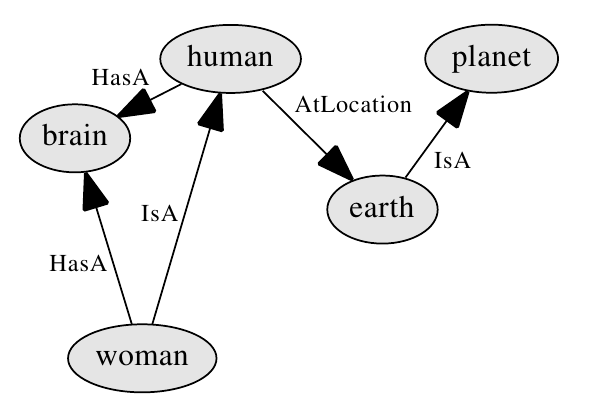}}
    \qquad
    \subfloat[]{\includegraphics[width=6cm]{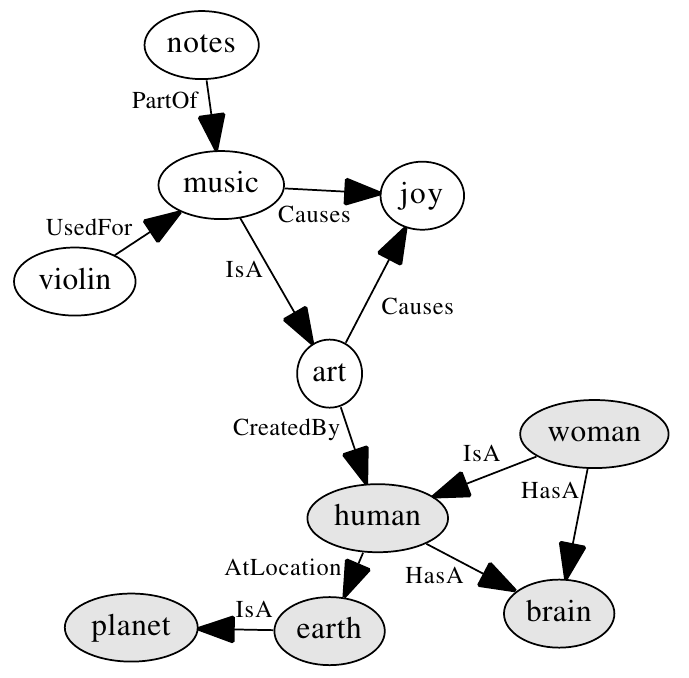}}
    \caption{Commonsense crossover type II (graph merging crossover). (a) Parent 1; (b) Parent 2; (c) Offspring, merging by the relation \texttt{CreatedBy(art, human)}. If no concepts attachable through commonsense relations are encountered, the offspring is formed by merging the parent networks as two separate clusters within the same semantic network.}
    \label{FigureCrossoverTypeII}
\end{figure*}

Given two parent networks, such as Figure~\ref{FigureCrossoverTypeII} (a) and (b), where no interchangeable concepts between these two can be located, the system falls back to the simpler type {II} crossover.

A concept from each parent that is \emph{attachable}\footnote{We define a distinct concept as \emph{attachable} to a semantic network if at least one commonsense relation connecting the concept to any of the concepts in the network can be discovered from commonsense knowledge bases.} to the other parent is selected as a \emph{crossover concept}.

The two parents are merged into an offspring by attaching a concept in one parent to another concept in the other parent, picked randomly out of all possible attachments (\texttt{CreatedBy(art, human)} in Figure~\ref{FigureCrossoverTypeII} (c). Another possibility is \texttt{Desires(human, joy)}.). The second offspring is formed randomly in the same way. In the case that no attachable concepts are found, the parents are merged as two separate clusters within the same individual.

\subsubsection{Commonsense mutation}

We introduce several types of \emph{commonsense mutation} operators that modify a parent by means of information from commonsense knowledge bases.

For each mutation to be performed, the type is picked at random with uniform probability. If the selected type of mutation is not feasible due to the commonsense structure of the parent, another type is again picked. In the case that a set timeout of $Count_{timeout}$ trials has been reached without any operation, the parent is returned as it is.

\subsubsection{Type I (Concept Attachment)}

A new concept randomly picked from the set of concepts \emph{attachable} to the parent is attached through a new relation to one of existing concepts (Figure~\ref{FigureMutation} (a) and (b)).

\subsubsection{Type IIa (Relation Addition)}

A new relation connecting two existing concepts in the parent is added, possibly connecting unconnected clusters within the same network (Figure~\ref{FigureMutation} (c) and (d)).

\subsubsection{Type IIb (Relation Deletion)}

A randomly picked relation in the parent is deleted, possibly leaving unconnected clusters within the same network (Figure~\ref{FigureMutation} (e) and (f)).

\subsubsection{Type IIIa (Concept Addition)}

A randomly picked new concept is added to the parent as a new cluster (Figure~\ref{FigureMutation} (g) and (h)).

\subsubsection{Type IIIb (Concept Deletion)}

A randomly picked concept is deleted with all the relations it is involved in, possibly leaving unconnected clusters within the same network (Figure~\ref{FigureMutation} (i) and (j)).

\subsubsection{Type IV (Concept Replacement)}

\begin{figure*}
    \centering
    \subfloat[]{\includegraphics[width=4cm]{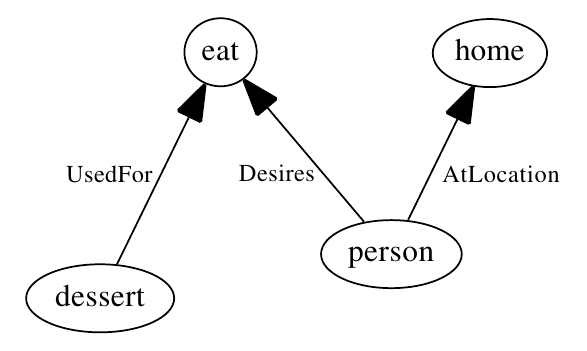}}
    \qquad
    \subfloat[]{\includegraphics[width=4cm]{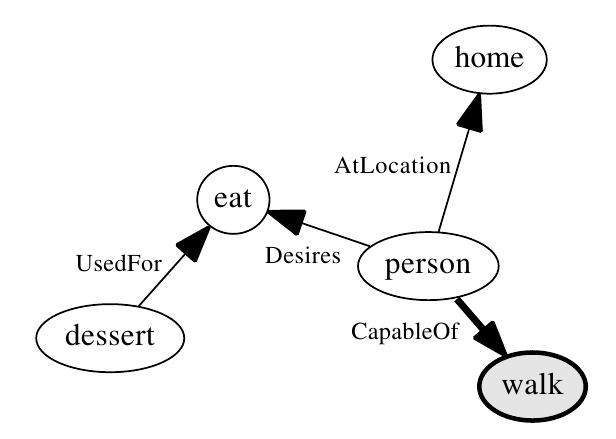}}
    \qquad
    \subfloat[]{\includegraphics[width=3cm]{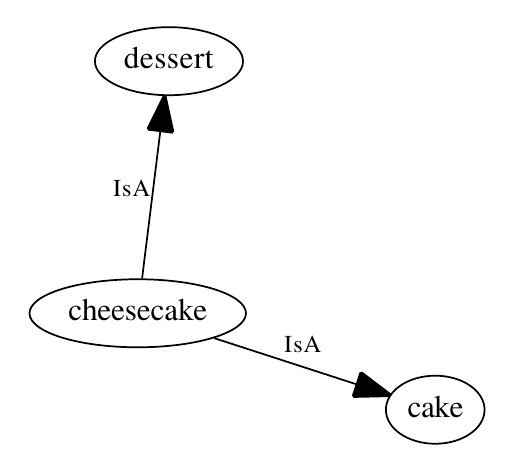}}
    \qquad
    \subfloat[]{\includegraphics[width=3cm]{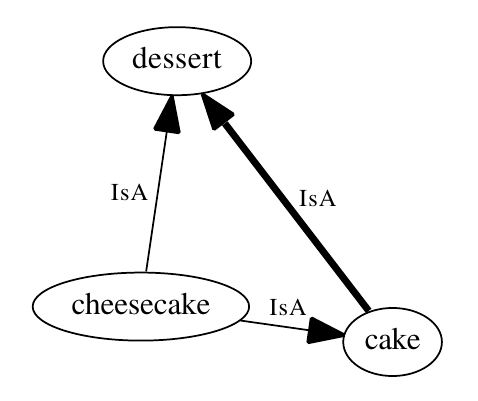}}
    \qquad
    \subfloat[]{\includegraphics[width=3cm]{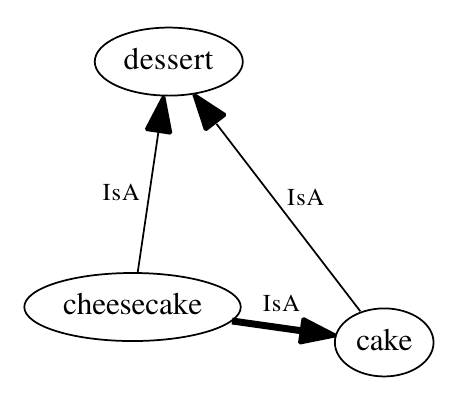}}
    \qquad
    \subfloat[]{\includegraphics[width=3cm]{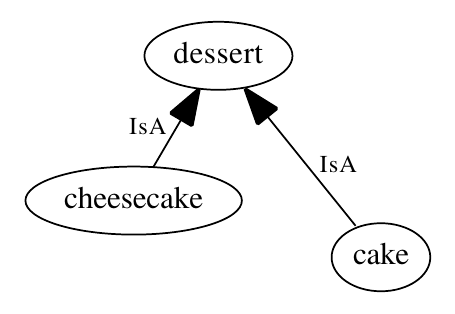}}
    \qquad
    \subfloat[]{\includegraphics[width=4cm]{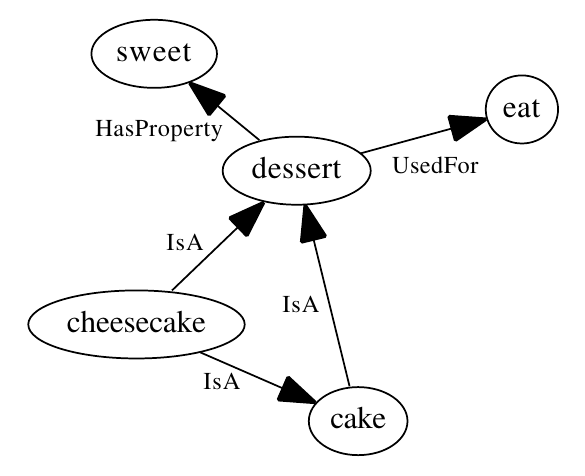}}
    \qquad
    \subfloat[]{\includegraphics[width=4cm]{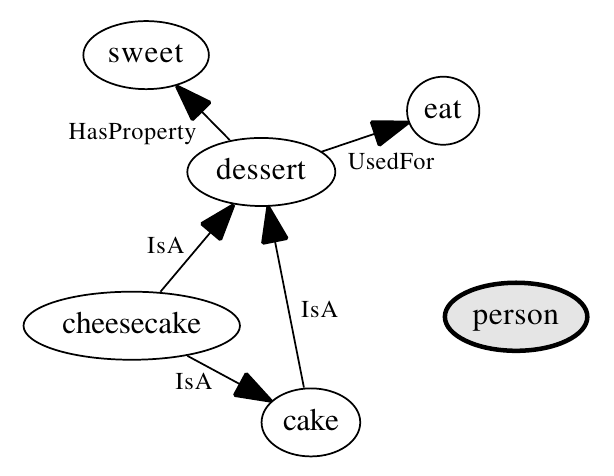}}
    \qquad
    \subfloat[]{\includegraphics[width=3.5cm]{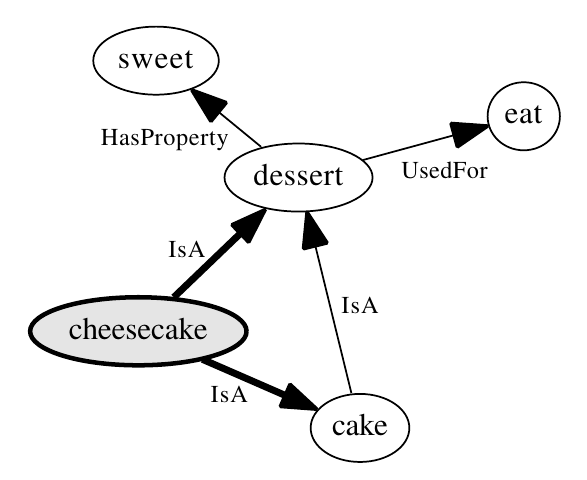}}
    \qquad
    \subfloat[]{\includegraphics[width=3.5cm]{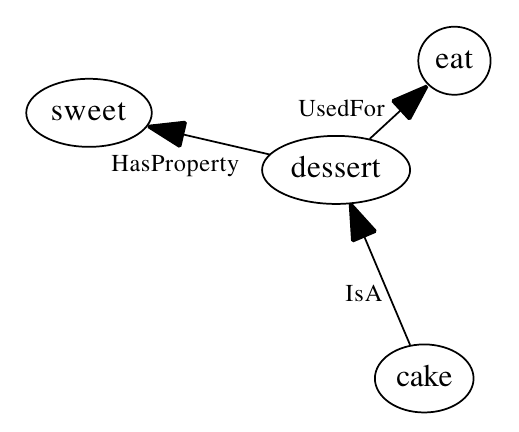}}
    \qquad
    \subfloat[]{\includegraphics[width=3.5cm]{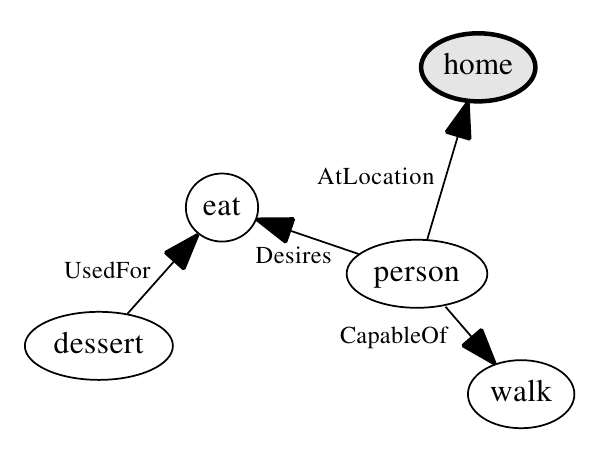}}
    \qquad
    \subfloat[]{\includegraphics[width=3.5cm]{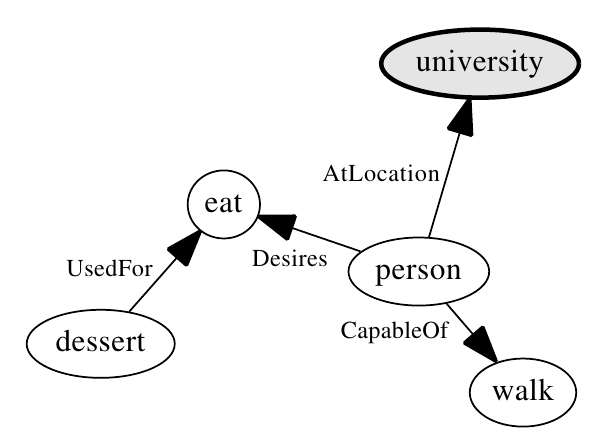}}
    \caption{Examples illustrating commonsense mutation. (a) Mutation type I (before); (b) Mutation type I (after); (c) Mutation type IIa (before); (d) Mutation type IIa (after); (e) Mutation type IIb (before); (f) Mutation type IIb (after); (g) Mutation type IIIa (before); (h) Mutation type IIIa (after); (i) Mutation type IIIb (before); (j) Mutation type IIIb (after); (k) Mutation type IV (before); (l) Mutation type IV (after).}
    \label{FigureMutation}
\end{figure*}

A concept in the parent, randomly picked from the set of those with at least one \emph{interchangeable} concept, is replaced with one of its interchangeable concepts, again randomly picked. Any relations left unsatisfied by the new concept are deleted (Figure~\ref{FigureMutation} (k) and (l)).

\section{Analogy as a fitness measure}
\label{SectionAnalogyAsAFitnessMeasure}

For experimenting with our approach, we select analogical reasoning as an initial application area, by using analogical similarity as our fitness measure.

This constitutes an interesting choice for evaluating our work, because it not only validates the viability of the novel algorithm, but also produces results of interest for the fields of analogical reasoning and computational creativity.

\subsection{Analogies and creativity}

There is evidence that analogical reasoning is at the core of higher-order cognition, and it enters into creative discovery, problem-solving, categorization, and learning \citep{Hofstadter2001}. Analogy-making ability is extensively linked with creative thought and regularly plays a role in creativity expressed in arts and sciences. \citet{Boden2009} classifies analogy as a form of \emph{combinational creativity}, noting that it works by producing unfamiliar combinations of familiar ideas.

In addition to literary use of metaphors and allegories in written language, analogies often constitute the basis of composition in all art forms including visual or musical. For example, in classical music, it is highly common to formulate interpretations of a composer's work in terms of tonal allegories \citep{Chafe1991}. In visual arts, examples of artistic analogy abound, ranging from allegorical compositions of Renaissance masters such as Albrect D\"{u}rer, to modern usage in film, such as the many layers of allegory in Stanley Kubrick's 2001: A Space Odyssey \citep{Pezzotta2012}.

In science, analogies have been used to convey revolutionary theories and models. A key example of analogy-based explanations is Kepler's explanation of the laws of heliocentric planetary motion with an analogy to light radiating from the Sun\footnote{Kepler argued, in his \emph{Astronomia Nova}, as light can travel undetectably on its way between the source and destination, and yet illuminate the destination, so can motive force be undetectable on its way from the Sun to planet, yet affect planet's motion.}.

\begin{figure*}
    \centering
    \includegraphics[width=0.8\textwidth]{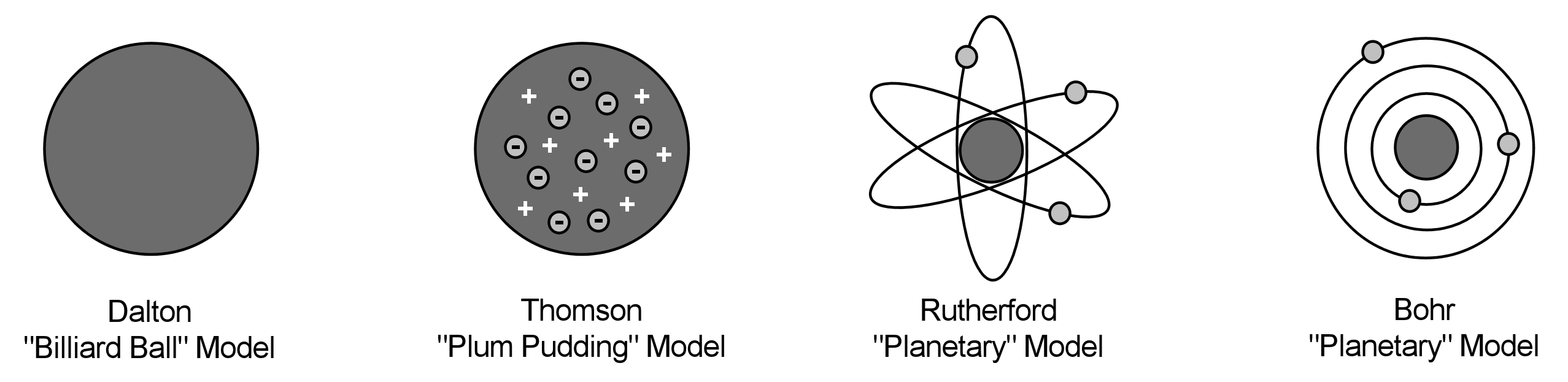}
    \caption{The Dalton (1805), Thomson (1904), Rutherford (1911), and Bohr (1913) models of the atom with their corresponding analogies.}
    \label{FigureAtomModels}
\end{figure*}

Another instance is Rutherford's analogy between the atom and the Solar System, where the internal structure of the atom is explained by electrons circling the nucleus in orbits like planets around the Sun. This model, which was later improved by Bohr to give rise to the Rutherford-Bohr model, was one of the ``planetary models'' of the atom, where the electromagnetic force between oppositely charged particles were presented analogous to the gravitational force between planetary bodies. Earlier models of the atom were, also notably, explained using analogies, including ``plum pudding'' model of Thomson and the ``billiard ball'' model of Dalton (Figure~\ref{FigureAtomModels}).

In contemporary studies, analogical reasoning is mostly seen through a structural point of view, framed by the structure mapping theory based on psychology \citep{Gentner1983}.

Other approaches to analogical reasoning include the view of \citet{Hofstadter1995} of analogy as a kind of high-level perception, where one situation is perceived as another one. \citet{Veale1997} extend the work in analogical reasoning to the more specific case of metaphors, which describe the understanding of one kind of thing in terms of another. A highly related cognitive theory is the \emph{conceptual blending} idea developed by \citet{Fauconnier2002}, which involves connecting several existing concepts to create new meaning, operating below the level of consciousness as a fundamental mechanism of cognition. An implementation of this idea is given by \citet{Pereira2007} as a computational model of abstract thought, creativity, and language.

\begin{figure}
    \centering
    \subfloat[]{\includegraphics[width=\columnwidth]{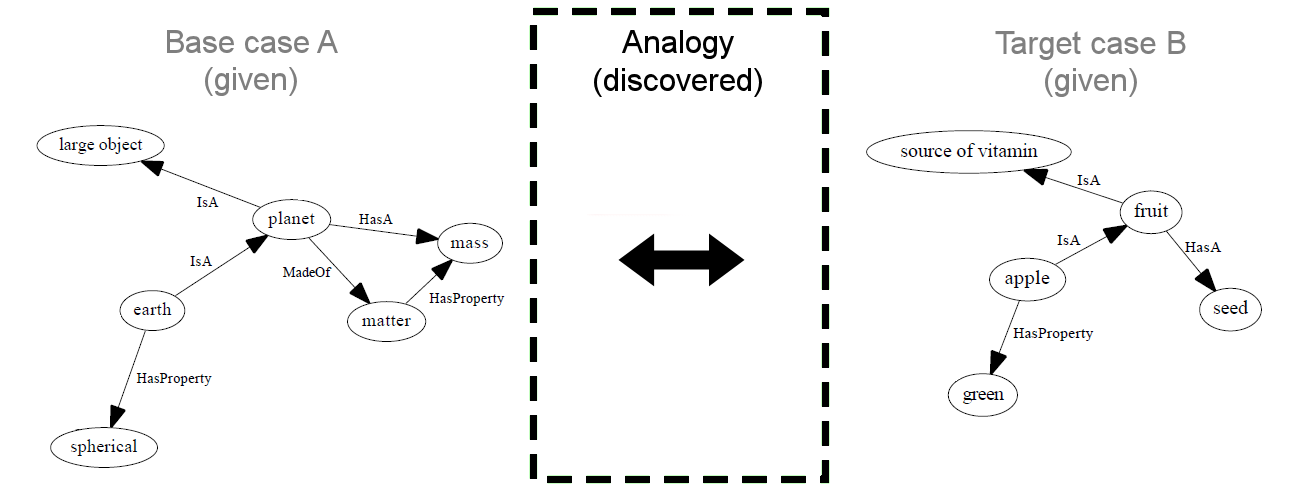}}
    \qquad
    \subfloat[]{\includegraphics[width=\columnwidth]{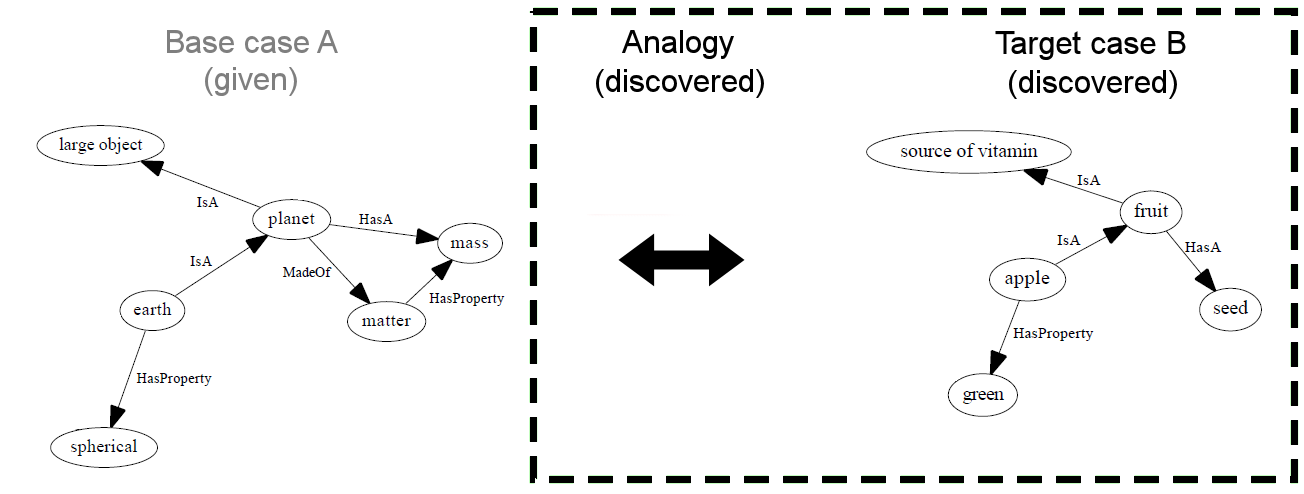}}
    \caption{Contribution to computational analogy-making. (a) Existing work in the field, restricted to finding analogical mappings between a given pair of domains (b) Our novel approach, capable of creating novel analogies as well as the analogous case itself.}
    \label{FigureAnalogyExistingAndNew}
\end{figure}

Computational approaches within the analogical reasoning field have been mostly concerned with the \emph{mapping} problem \citep{French2002}. Put in a different way, models developed and implemented are focused on constructing mappings between two given source and target domains (Figure~\ref{FigureAnalogyExistingAndNew} (a)). This focus neglects the problem of retrieval or recognition of a new source domain, given a target domain, or the other way round.

By combining our algorithm for the evolution of semantic networks with a fitness measure based on analogical similarity, we can essentially produce a method to address this creativity-related subproblem of analogical reasoning, which has remained, so far, virtually untouched.

We accomplish this by

\begin{enumerate}
    \item providing our evolutionary algorithm with a ``reference'' semantic network that will represent the input to the system; and
    \item running the evolutionary process under a fitness measure quantifying analogical similarity to the given ``reference'' network.
\end{enumerate}

This, in effect, creates a ``survival of the fittest analogies'' process where, starting from a random initial population of semantic networks, one gets semantic networks that get gradually more analogous to the given reference network.

In our implementation, we define the fitness measure to take the reference semantic network as the \emph{base} and the individual whose fitness is just being evaluated as the \emph{target}. In other terms, this means that the system produces structurally analogous target networks for a given base network. From a computational creativity perspective, an interpretation for this would be the ``imagining'', or creation, of a novel case that is analogous to a case at hand.

This designation of the base and target roles for the two networks is an arbitrary choice, and it is straightforward to define the fitness function in the other direction. So, if the system would be set up such that it would produce base networks, given the target network, one can then interpret this as the the classical retrieval process in analogical reasoning, where one is supposed to retrieve a base case that is analogous to the currently encountered case, for using it as a basis for solution.

If one subscribes to the ``retrieval of a base case'' interpretation, since the ultimate source of all the information underlying the generated networks is the commonsense knowledge bases, one can treat this source of knowledge as a part of the system's memory, and see it as a ``generic case base'' from which the base cases are retrieved.

On the other hand, if we consider the ``imagination of a novel case'' interpretation, our system, in fact, replicates a mode of behavior observed in psychology research where an analogy is not always simply \emph{recognized} between an original case and a retrieved analogous case from memory, but the analogous case can sometimes be \emph{created} together with the analogy \citep{Clement1988}.

Considering the depth of commonsense knowledge sources, this creation process is effectively open-ended; and due to randomly performed queries, it produces different analogous cases in each run. This capability of open-ended creation of novel analogous cases is, to our knowledge, the first of its kind and makes our approach interesting for the analogical reasoning and computational creativity fields. 

The random nature of population initialization and the breadth of information in ConceptNet and WordNet virtually ensure that the generated semantic networks are in different domains from the one supplied as the input. However, it is possible to formulate fitness measures that include a measure of \emph{semantic similarity}\footnote{Readily available by using WordNet \cite{Pedersen2004}.} in addition to analogical similarity and to penalize networks that are semantically too similar to the source network.

\subsection{Structure mapping engine}

The Structure Mapping Engine (SME) \citep{Falkenhainer1989} is an analogical matching algorithm firmly based on the psychological \emph{structure mapping theory} of Gentner. It is a very robust algorithm, having been used in many practical applications by a variety of research groups, and it has been considered the most influential work on the modeling of analogy-making \citep{French2002}.

An important characteristic of SME is that it ignores surface features and it can uncover mappings between potentially very distant domains, if they have a similar representational structure.

\begin{figure}
    \centering
    \subfloat[]{\includegraphics[width=\columnwidth]{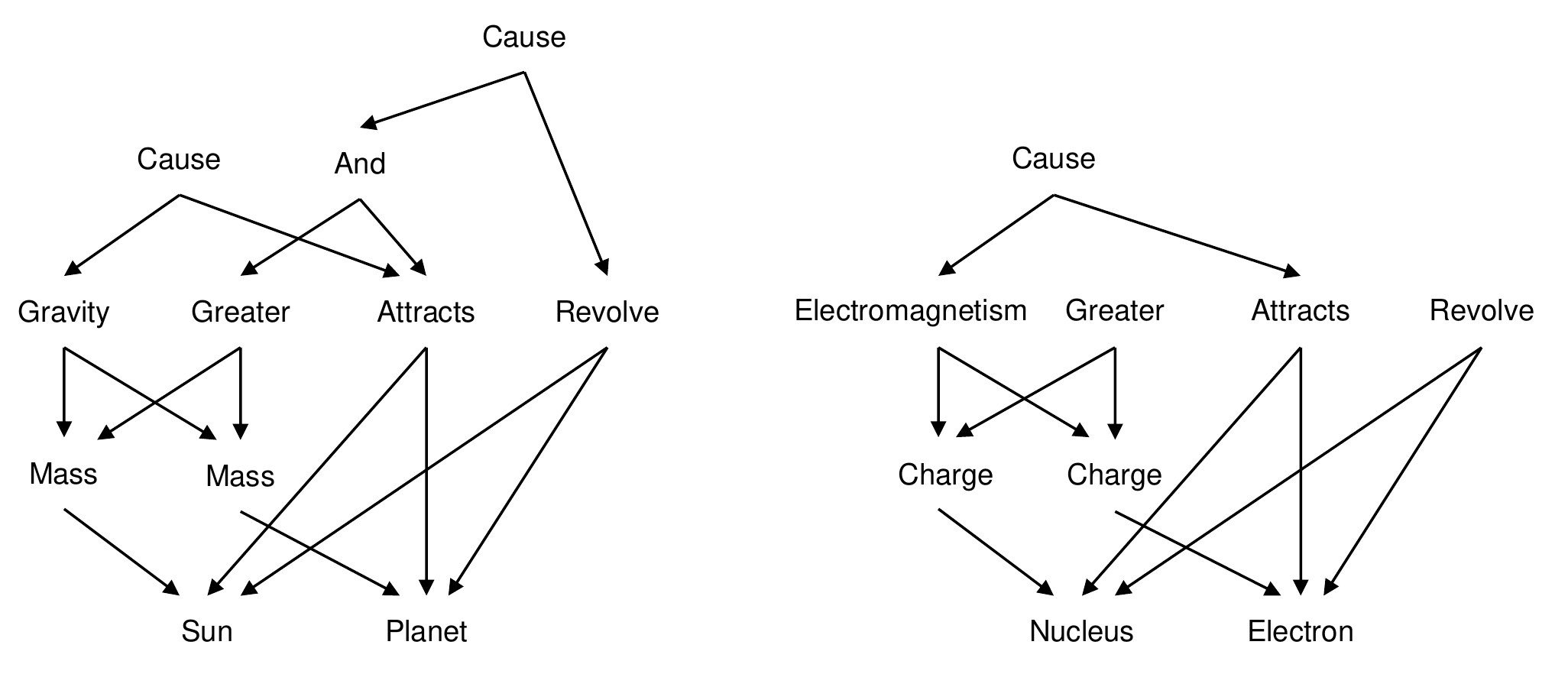}}
    \qquad
    \subfloat[]{\includegraphics[width=\columnwidth]{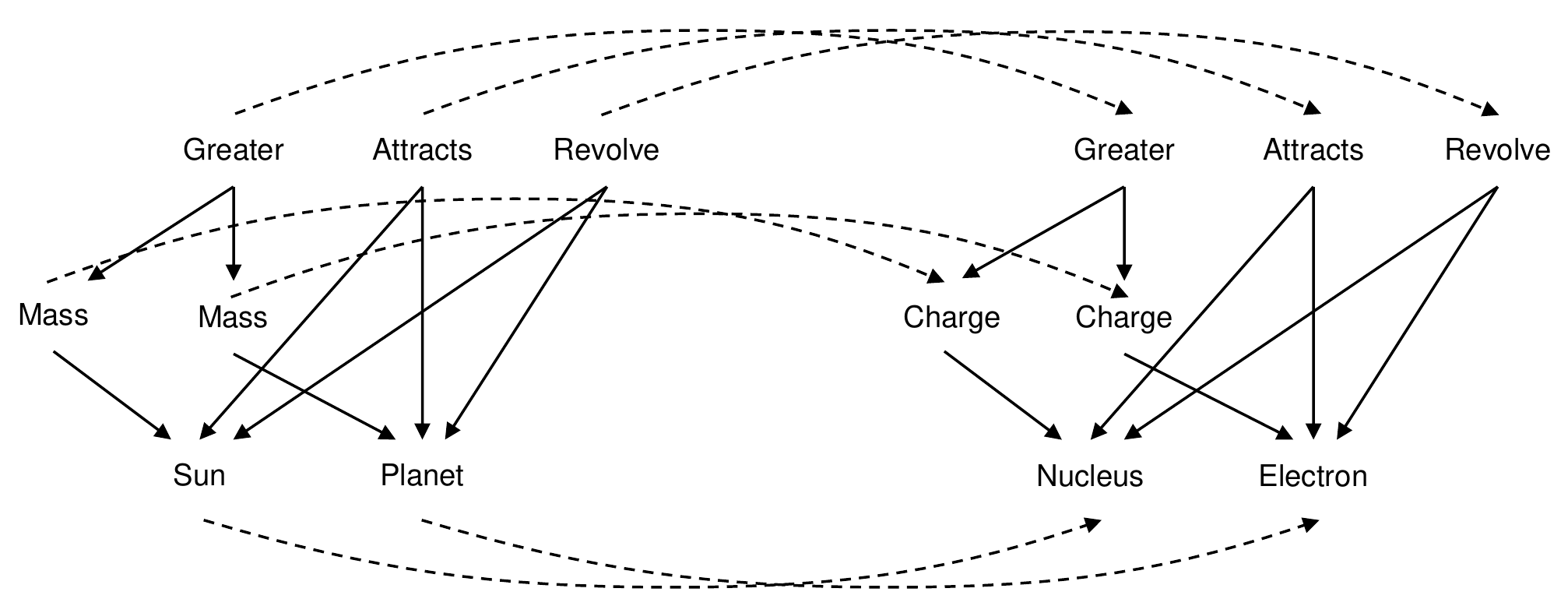}}
    \caption{Representation of the Rutherford-Bohr atom model -- Solar System analogy as graphs: (a) Predicates about the two domains. (b) Analogy as a mapping of structure between the two domains.}
    \label{FigureStructureMappingEngineAtomSolarSystem}
\end{figure}

A typical example given for illustrating the working of SME is the analogy between the Rutherford-Bohr atom model and the Solar System, which we already mentioned. Using a predicate calculus representation, Figure~\ref{FigureStructureMappingEngineAtomSolarSystem} illustrates a structural mapping between these domains.

Here we make use of our own implementation of SME based on the original description by \citet{Falkenhainer1989} and adapt it to the simple structure of semantic networks. 

Using SME in this way necessitates the introduction of a mapping between the concept--relation based structure of semantic networks and the predicate calculus based representation traditionally used in SME applications.

A highly versatile such mapping is given by \citet{Larkey2003}. Given information such as ``Jim (a man) loves Betty (a woman)'', one can transform the predicate calculus representation of \texttt{loves(Jim, Betty)}, \texttt{gender(Jim, male)}, \texttt{gender(Betty, female)} into a semantic network representation by converting predicates into nodes such as \texttt{gender} and \texttt{loves}; and creating argument nodes for each argument of a predicate. This kind of mapping makes it possible, theoretically, to represent arbitrarily complex information within the simple representation framework of semantic networks. As an example, one can represent meta-information such as ``John knows that Jim loves Betty''.

However, the approach of \citet{Larkey2003} requires the creation of ad hoc ``relation nodes'' for the representation of relations between concepts and the usage of unlabeled directed edges. On the other hand, the existing structure of the commonsense knowledge bases that we interface extensively, mainly ConceptNet, are based on nodes representing concepts and labeled directed edges representing relations. In this representation, nodes can have arbitrary names but the names of edges come from a limited set of basic relation names\footnote{For ConceptNet version 4: \texttt{IsA}, \texttt{HasA}, \texttt{PartOf}, \texttt{UsedFor}, \texttt{AtLocation}, \texttt{CapableOf}, \texttt{MadeOf}, \texttt{CreatedBy}, \texttt{HasSubevent}, \texttt{HasFirstSubevent}, \texttt{HasLastSubevent}, \texttt{HasPrerequisite}, \texttt{MotivatedByGoal}, \texttt{Causes}, \texttt{Desires}, \texttt{CausesDesire}, \texttt{HasProperty}, \texttt{ReceivesAction}, \texttt{DefinedAs}, \texttt{SymbolOf}, \texttt{LocatedNear}, \texttt{ObstructedBy}, \texttt{ConceptuallyRelatedTo}, \texttt{InheritsFrom}.}. Because of this, we take another approach for mapping between semantic networks and predicate calculus.

Due to these reasons, we define a basic list of correspondences between the two representation schemes, where we treat ``entities'' as concepts, relations as relations, attributes as \texttt{IsA} relations; and exclude functions. Table~\ref{TableSMESemanticNetworks} gives the list of correspondences that we employ in our SME implementation.

\begin{table}
  \caption{Correspondences between SME predicate calculus statements \citep{Falkenhainer1989} and semantic network structure that we define for applying structure mapping to semantic networks.}
  \label{TableSMESemanticNetworks}
  \begin{tabularx}{\columnwidth}{@{}Xl@{}} 
    \toprule
    Predicate calculus & Semantic networks\\
    \midrule
    Entity & Concept (node)\\
    Relation & Relation (edge)\\
    Attribute & \texttt{IsA} or \texttt{HasProperty} relation\\
    Function & Not employed\\
    \bottomrule
  \end{tabularx}
\end{table}

\subsection{Results}

With the implementation of analogical similarity-based fitness measure that we described so far, we carried out numerous experiments with reference networks representing different domains. In this part, we present the results from two such experiments.

Table~\ref{TableNewMAAnalogyParameters} provides an overview of the parameter values that we used for conducting these experiments.

The selection of crossover and mutation probabilities for a particular application have been a traditional subject of debate in EA literature \citep{Spears1992}. Since the foundation of the field, in essence, the arguments have been mainly centered on the relative importance of the crossover and mutation operators in the progress of evolution. For our approach, we make the decision to follow the somewhat established consensus in the graph-based EA field \citep{Pereira1999}, dominated by genetic programming (GP) and the selection of parameters by the pioneering work of Koza. 

Thus, we use a crossover probability of $Prob_{rec} = 0.85$, similar to the high crossover probabilities typically $\geq 0.9$ encountered in GP literature \citep{Koza2003}. 

However, unlike the typical GP mutation value of $\leq 0.1$, we employ a somewhat-above-average mutation rate of $Prob_{mut} = 0.15$.

Due to the fact that our algorithm is the first attempt at having a graph-based evolutionary model of memetics, this mutation rate is somewhat arbitrary and is dependent on our subjective interpretation of the mutation events in memetic processes. Nonetheless, there is preliminary support for a high mutation rate in memetics, where it has been postulated, for example by \citet{GilWhite2008}, that memes would have a high tendency of mutation.

We select a population size of $Size_{pop} = 200$ individuals, and subject this population to tournament selection with a tournament size of $Size_{tourn} = 8$ and a winning probability $Prob_{win} = 0.8$.

\begin{table}
    \caption{Parameter set used during experiments. Refer to Table~\ref{TableNewMAParameters} for an explanation of parameters.}
    \label{TableNewMAAnalogyParameters}
    \begin{tabularx}{\columnwidth}{@{}lXX@{}} 
        \toprule
        & Parameter & Value\\
        \midrule
        Evolution & $Size_{pop}$ & 200\\
        & $Prob_{rec}$ & 0.85\\
        & $Prob_{mut}$ & 0.15\\
        \midrule
        Semantic networks & $Size_{network}$ & 5\\
        & $Score_{min}$ & 2\\
        & $Count_{timeout}$ & 10\\
        \midrule
        Tournament selection & $Size_{tourn}$ & 8\\
        & $Prob_{win}$ & 0.8\\
        \bottomrule
    \end{tabularx}
\end{table}

\begin{figure}
  \centering
  \includegraphics[width=\columnwidth]{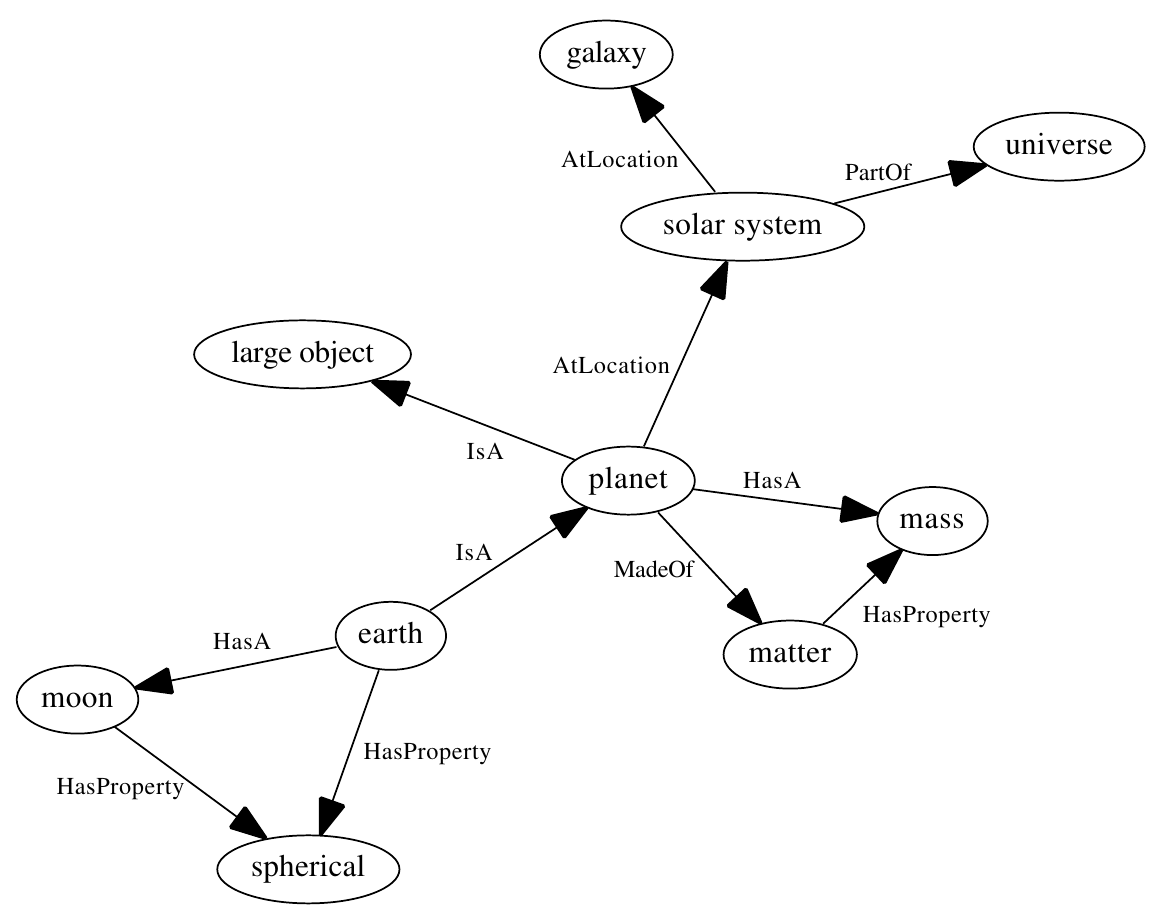}
  \caption{Experiment 1: Given semantic network, 10 concepts, 11 relations (base domain).}
  \label{FigureNewMAAnalogyExperiment1Base}
\end{figure}

\begin{figure}
  \centering
  \includegraphics[width=\columnwidth]{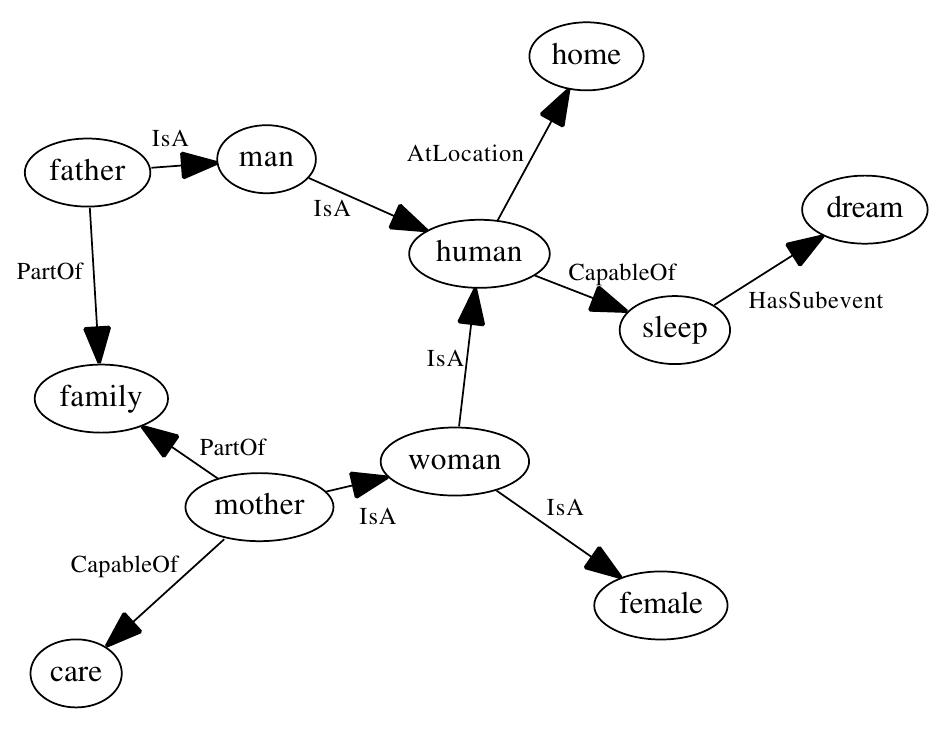}
  \caption{Experiment 2: Given semantic network, 11 concepts, 11 relations (base domain).}
  \label{FigureNewMAAnalogyExperiment2Base}
\end{figure}

Using this parameter set, here we present the results from two runs of experiment: 

\begin{enumerate}
    \item analogies generated for a network describing some basic astronomical knowledge, shown in Figure~\ref{FigureNewMAAnalogyExperiment1Base}; and
    \item analogies generated for a network describing familial relations, shown in Figure~\ref{FigureNewMAAnalogyExperiment2Base}.
\end{enumerate}

\begin{figure}
  \centering
  \includegraphics[width=\columnwidth]{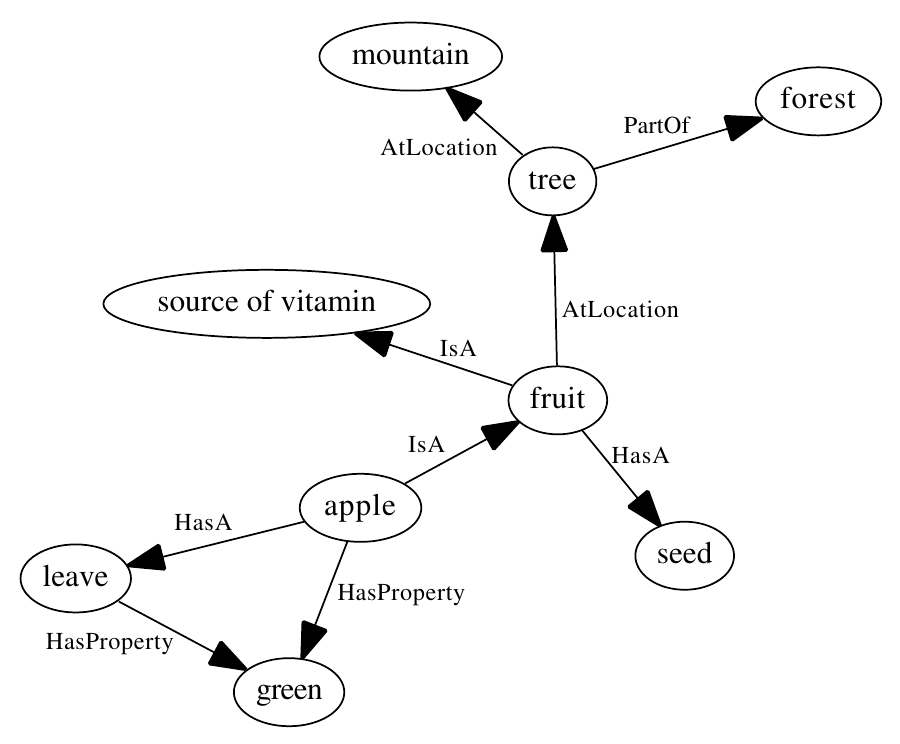}
  \caption{Experiment 1: Evolved individual, 9 concepts, 9 relations (target domain). The evolved individual is encountered after 35 generations, with fitness value 2.8. Concepts and relations of the individual not involved in the analogy are not shown here for clarity.}
  \label{FigureNewMAAnalogyExperiment1Target}
\end{figure}

For the first reference base network (Figure~\ref{FigureNewMAAnalogyExperiment1Base}), after a run of the algorithm for 35 generations, the system produced the target network shown in Figure~\ref{FigureNewMAAnalogyExperiment1Target}.

The produced target network exhibits an almost one-to-one structural correspondence with the reference network, missing only one node (\texttt{mass} in the original network) and two relations both pertaining to this missing node (\texttt{HasA(planet, mass)} and \texttt{HasProperty(matter, mass)}). The discovered analogy is remarkably inventive, and draws a parallel between the Earth and an apple: Just as the Earth is like an apple, planets are like fruits and the solar system is like a tree holding these fruits. Just as the solar system is a part of the universe, a tree is a part of a forest.

It is an intuitive analogy and leaves us with the impression that it is comparable with the classic analogy between the atom and the Solar System that we mentioned in the beginning of this section. Table~\ref{TableNewMAAnalogyExperiment1} gives a full list of all the correspondences.

\begin{figure}
  \centering
  \includegraphics[width=\columnwidth]{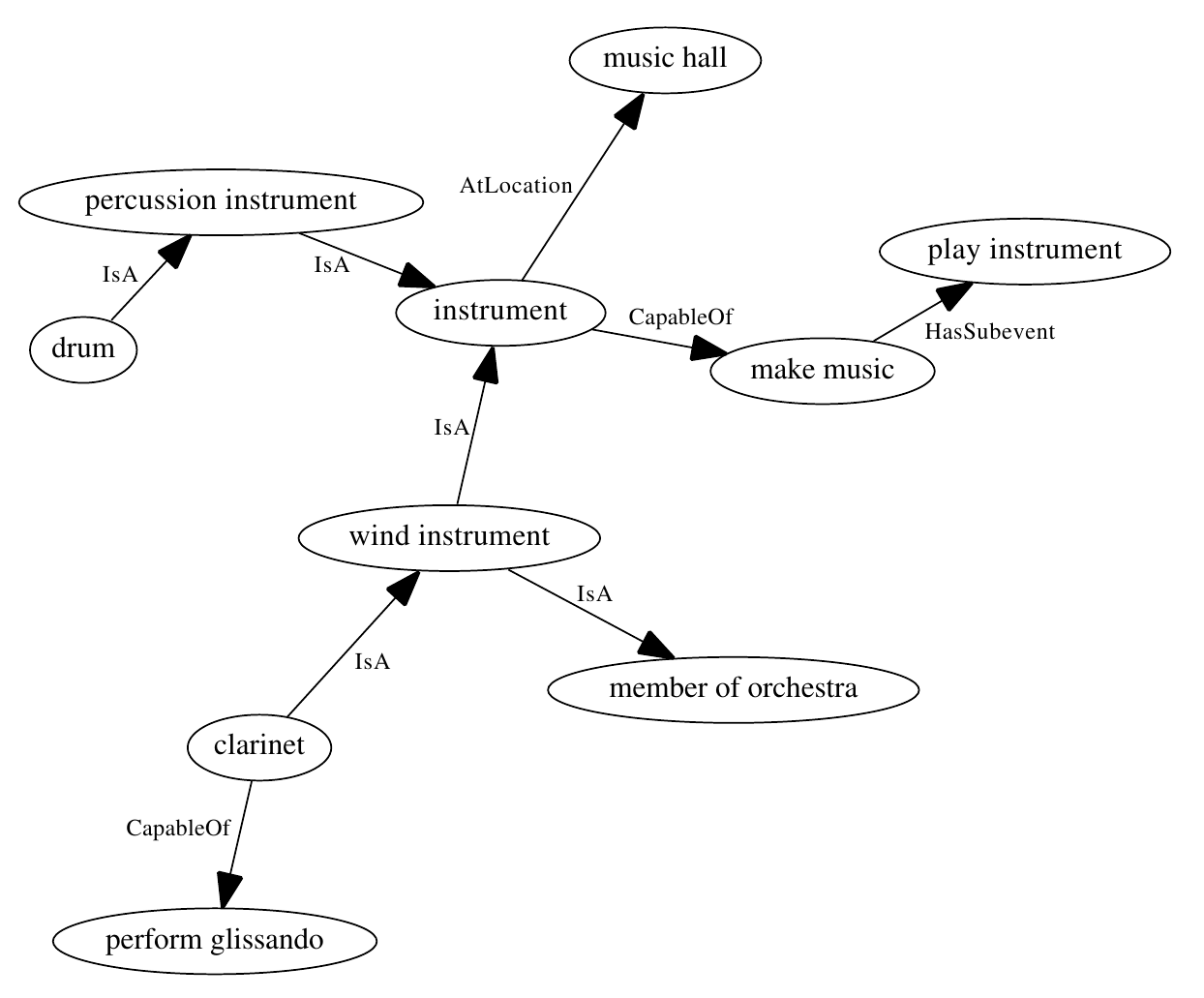}
  \caption{Experiment 1: Evolved individual, 10 concepts, 9 relations (target domain). The evolved individual is encountered after 42 generations, with fitness value 2.7. Concepts and relations of the individual not involved in the analogy are not shown here for clarity.}
  \label{FigureNewMAAnalogyExperiment2Target}
\end{figure}

For the second reference network (Figure~\ref{FigureNewMAAnalogyExperiment2Base}), in a run after 42 generations, our algorithm produced the network shown in Figure~\ref{FigureNewMAAnalogyExperiment2Target}.

The produced analogy can be again considered ``creative'', drawing a parallel between human beings and musical instruments. It considers a mother as a clarinet and a father as a drum; and just as a mother is a woman and a father a man, a clarinet is an instance of wind instrument and a drum is an instance of percussion instrument. The rest of the correspondences also follow in a somewhat intuitive way. Again, Table~\ref{TableNewMAAnalogyExperiment2} gives a list of correspondences.

We should note here that each of these two examples were hand-picked out of a collection of approximately hundred runs with the corresponding reference network, chosen because they represent interesting analogies suggesting possible creative value. It is evidently a subjective judgment of what would be ``interesting'' to present to our audience. This is a common issue in computational creativity research, recognized, for example, by \citet{Colton2012} who introduce the term \emph{curation coefficient} as an informal subjective measure of typicality, novelty, and quality of the output from generative algorithms.

\begin{table}
    \caption{Experiment 1: Correspondences between the base and target networks, after 35 generations.}
    \label{TableNewMAAnalogyExperiment1}
    \begin{tabularx}{\columnwidth}{@{}XX@{}}
        \toprule
        Base & Target\\
        \midrule
        \textbf{Concepts} & \\
        \texttt{earth} & \texttt{apple}\\
        \texttt{moon} & \texttt{leave}\\
        \texttt{planet} & \texttt{fruit}\\
        \texttt{solar system} & \texttt{tree}\\
        \texttt{galaxy} & \texttt{forest}\\
        \texttt{universe} & \texttt{forest}\\
        \texttt{spherical} & \texttt{green}\\
        \texttt{matter} & ---\\
        \texttt{mass} & \texttt{seed}\\
        \texttt{large object} & \texttt{source of vitamin}\\
        \midrule
        \textbf{Relations} & \\
        \texttt{HasA(earth, moon)} & \texttt{HasA(apple, leave)}\\
        \texttt{HasProperty(earth, spherical)} & \texttt{HasProperty(apple, green)}\\
        \texttt{HasProperty(moon, spherical)} & \texttt{HasProperty(leave, green)}\\
        \texttt{IsA(earth, planet)} & \texttt{IsA(apple, fruit)}\\
        \texttt{IsA(planet, large object)} & \texttt{IsA(fruit, source of vitamin)}\\
        \texttt{AtLocation(planet, solar system)} & \texttt{AtLocation(fruit, tree)}\\
        \texttt{AtLocation(solar system, galaxy)} & \texttt{AtLocation(tree, mountain)}\\
        \texttt{PartOf(solar system, universe)} & \texttt{PartOf(tree, forest)}\\
        \texttt{MadeOf(planet, matter)} & ---\\
        \texttt{HasA(planet, mass)} & \texttt{HasA(fruit, seed)}\\
        \texttt{HasProperty(matter, mass)} & ---\\
        \bottomrule
    \end{tabularx}
\end{table}

\begin{table}
    \caption{Experiment 2: Correspondences between the base and target networks, after 42 generations.}
    \label{TableNewMAAnalogyExperiment2}
    \begin{tabularx}{\columnwidth}{@{}XX@{}}
        \toprule
        Base & Target\\
        \midrule
        \textbf{Concepts} & \\
        \texttt{mother} & \texttt{clarinet}\\
        \texttt{father} & \texttt{drum}\\
        \texttt{woman} & \texttt{wind instrument}\\
        \texttt{man} & \texttt{percussion instrument}\\
        \texttt{human} & \texttt{instrument}\\
        \texttt{home} & \texttt{music hall}\\
        \texttt{care} & \texttt{perform glissando}\\
        \texttt{family} & ---\\
        \texttt{sleep} & \texttt{make music}\\
        \texttt{dream} & \texttt{play instrument}\\
        \texttt{female} & \texttt{member of orchestra}\\
        \midrule
        \textbf{Relations} & \\
        \texttt{IsA(mother, woman)} & \texttt{IsA(clarinet, wind instrument)}\\
        \texttt{IsA(father, man)} & \texttt{IsA(drum, percussion instrument)}\\
        \texttt{IsA(woman, human)} & \texttt{IsA(wind instrument, instrument)}\\
        \texttt{AtLocation(human, home)} & \texttt{AtLocation(instrument, music hall)}\\
        \texttt{IsA(man, human)} & \texttt{IsA(percussion instrument, instrument)}\\
        \texttt{PartOf(mother, family)} & ---\\
        \texttt{PartOf(father, family)} & ---\\
        \texttt{CapableOf(mother, care)} & \texttt{CapableOf(clarinet, perform glissando)}\\
        \texttt{CapableOf(human, sleep)} & \texttt{CapableOf(instrument, make music)}\\
        \texttt{HasSubevent(sleep, dream)} & \texttt{HasSubevent(make music, play instrument)}\\
        \texttt{IsA(woman, female)} & \texttt{IsA(wind instrument, member of orchestra)}\\
        \bottomrule
    \end{tabularx}
\end{table}

During our experiments, we observed that under the selected parameter set, the evolutionary process approaches equilibrium conditions after approximately 50 generations. This behavior is typical and expected in EA approaches and manifests itself with an initial exponential or logarithmic growth in fitness that asymptotically approaches a fitness plateau, after which fitness increasing events will be sporadic and negligible.

\begin{figure}
  \centering
  \includegraphics[width=\columnwidth]{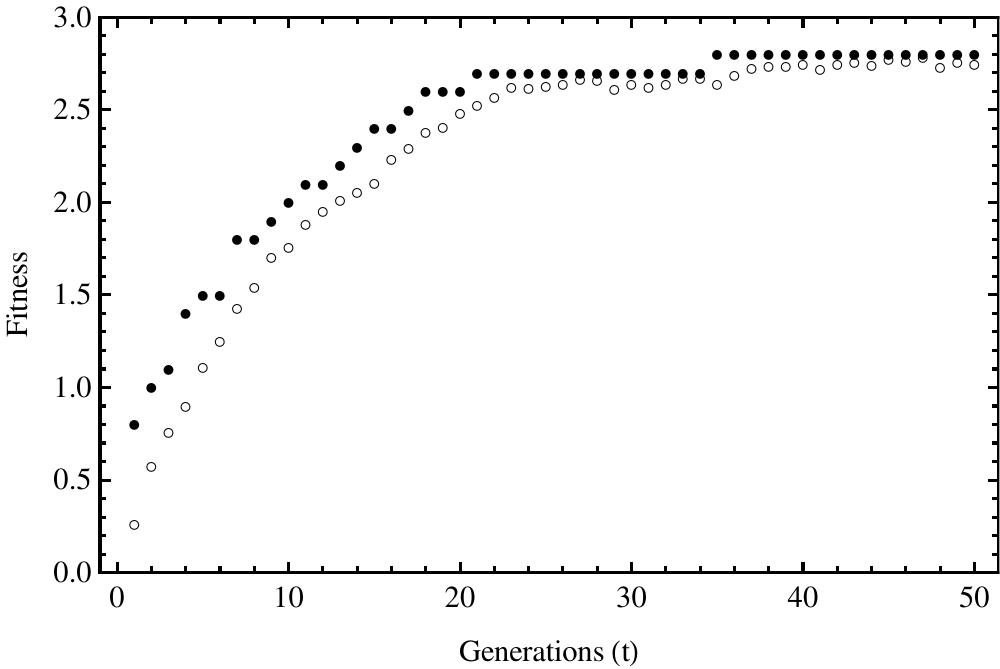}
  \caption{Progress of fitness during a typical run with parameters given in Table~\ref{TableNewMAAnalogyParameters}. Filled circles represent the best individual in a generation, while the empty circles represent population average.}
  \label{FigureNewMAAnalogyFitness}
\end{figure}

Figure~\ref{FigureNewMAAnalogyFitness} shows the progression of the average fitness of the population and the fitness of the best individual for each passing generation, during the course of one of our experiments with the reference network in Figure~\ref{FigureNewMAAnalogyExperiment1Base}, which lasted for 50 generations. We observe that the evolution process asymptotically reaches a fitness plateau after about 40 generations.

\begin{figure}
  \centering
  \includegraphics[width=\columnwidth]{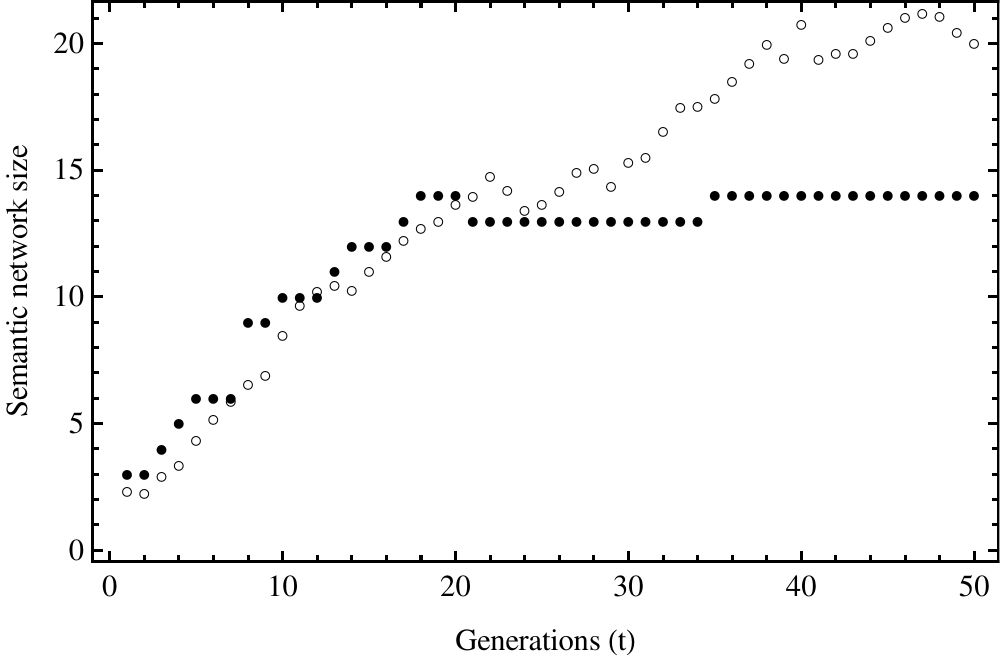}
  \caption{Progress of semantic network size during a typical run with parameters given in Table~\ref{TableNewMAAnalogyParameters}. Filled circles represent the best individual in a generation, while the empty circles represent population average. Network size is taken to be the number of relations (edges) in the semantic network.}
  \label{FigureNewMAAnalogyNetworkSize}
\end{figure}

Coinciding with the progression of fitness values, we observe, in Figure~\ref{FigureNewMAAnalogyNetworkSize}, the sizes of individual semantic networks both for the best individual and as a population average. Just as in the fitness values, there is a pronounced stabilization of the network size for the best individual in the population, occurring around the 40th generation. While the value stabilizes for the best individual, the population average for the network size keeps a trend of (gradually slowing) increase.

Our interpretation of this phenomenon is that, once the size of the best network becomes comparable with the size of the given reference network (Figure~\ref{FigureNewMAAnalogyExperiment1Base}, comprising 10 concepts and 11 relations) and the analogies considered by the SME algorithm have already reached a certain quality, further increases in the network size would not cause substantial improvement on the SME structural evaluation score. This is because the analogical mapping from the reference semantic network to the current best individual is already highly optimized and very close to the ideal case of a structurally one-to-one mapping (cf. Figure~\ref{FigureNewMAAnalogyExperiment1Base}, 10 concepts, 11 relations, and Figure~\ref{FigureNewMAAnalogyExperiment1Target}, 9 concepts, 9 relations).

In general, our experiments demonstrate that, combined with the SME-based fitness measure, the algorithm we developed is capable of spontaneously creating collections of semantic networks analogous to the one given as reference. In most cases, our implementation was able to reach extensive analogies within 50 generations and reasonable computational resources, where a typical run of experiment took around 45 minutes on a medium-range laptop computer with AMD Athlon II 2.2 GHz processor and 8 GB of RAM.

\section{Conclusions}
\label{SectionConclusions}

We presented a novel graph-based EA employing semantic networks as evolving individuals. The use of semantic networks provides a simple yet powerful means of representing pieces of evolving knowledge, giving us a possibility to interpret this algorithm as an implementation of the idea of memetics. Because this work constitutes a novel semantic network-based EA, we had to establish the necessary crossover and mutation operators working on this representation.

We make extensive use of commonsense reasoning and commonsense knowledge bases, necessitated by the semantic network-based representation and the requirement that all operations should ensure meaningful conceptual relations. Put another way, we use a combination of random processes constrained by the non-random structural bounds of commonsense knowledge, under selection pressure of the defined fitness function.

For evaluating the approach, we make use of SME as the basis of a fitness function that measures analogical similarity. With the analogical similarity-based fitness calculated between the reference network and the evolving networks in the population, we create a system capable of spontaneously generating networks analogous to any given network. This system represents a first in the analogical reasoning field, because current models have been limited to only finding analogical mappings between two already existing networks.

\subsection{Limitations and future work}

The most considerable limitation of this work comes from our choice of using semantic networks instead of a more powerful representation scheme. For example, since we are using SME for experimenting with our approach, it would be highly desirable and logical to use predicate calculus to represent evolving individuals. Instead, we limit the representation to semantic networks, and provide our own implementation of SME that we adapt to work on the simple directed graph structure of semantic networks.

This choice of limiting representation was mainly directed by our reliance on ConceptNet version 4 as the main commonsense knowledge base used in this study, which is based on simple binary relations using a limited set of relation types. This impedes the representation of more complex information such as temporal relations or causal connections between subgraphs. It should be noted, however, that in the next version, ConceptNet project has made a decision to move to a ``hypergraph'' representation, where one can have relations about other instances of relation between concepts. This can, in effect, greatly increase the expressivity of the system.

Another issue in the current study is the selection of parameter values for our EA implementation. Due to the fact that our algorithm is a first attempt at having a graph-based implementation of memetics, we are faced with selecting mutation and crossover rates without any antecedents. Even in theoretical studies of cultural evolution, discussions of the frequency of variation events are virtually nonexistent. This makes our parameter values rather arbitrary, roughly guided by the general conventions in the graph-based EA field.

For future work, it would be interesting to experiment with extensions of the simple SME-based fitness measure that we have used. As semantic networks are graphs, a straightforward possibility is to take graph-theoretical properties of candidate networks into account, such as the clustering coefficient or shortest path length. With these kinds of constraints, selection pressure on the network structure can be adjusted in a more controlled way.

Another highly interesting prospect with the EA system would be to consider different types of mutation and crossover operators, and doing the necessary study for grounding the design of such operators on existing theories of cultural transmission and variation. Combined with realistically formed fitness functions, one can use such a system for modeling selectionist theories of knowledge. Performing experiments with such a setup could be considered a ``memetic simulation'' and comparable to computational simulations of genetic processes performed in computational biology.

Besides the ``memetic'' interpretation, a more hands-on application that we foresee we can achieve in the short-term is practical computational creativity. Already with the SME-based fitness measure that we demonstrated in this article, it would be possible to create systems for tasks such as story generation based on analogies \citep{Zhu2013}. This would involve giving the system an existing story as the input, and getting an analogous story in another domain as the output. For doing this we would need to define a structural representation scheme of story elements, and, preferably an automated way of translating between structural and textual representations.

\begin{acknowledgements}
This work was supported by a JAE-Predoc fellowship from CSIC, and the research grants: 2009-SGR-1434 from the Generalitat de Catalunya, CSD2007-0022 from MICINN, and Next-CBR TIN2009-13692-C03-01 from MICINN. We thank the three anonymous reviewers whose input has considerably improved the article.
\end{acknowledgements}

%\bibliographystyle{plainnat}
\bibliographystyle{spmpscinat}
\bibliography{BaydinLopezDeMantarasOntanon}   % name your BibTeX data base

\end{document}